\documentclass[TRO]{RLIpaper}

\usepackage{geometric_algebra}

\begin{document}
\title{Geometric Algebra for Optimal Control with Applications in Manipulation Tasks}
\author{Tobias Löw and Sylvain Calinon
    \thanks{The authors are with the Idiap Research Institute, Martigny, Switzerland and with EPFL, Lausanne, Switzerland. (\tt\footnotesize tobias.loew@idiap.ch)}
    \thanks{This work was supported by the State Secretariat for Education, Research and Innovation in Switzerland for participation in the European Commission’s Horizon Europe Program through the INTELLIMAN project (https://intelliman-project.eu/, HORIZON-CL4-Digital-Emerging Grant 101070136) and the SESTOSENSO project (http://sestosenso.eu/, HORIZON-CL4-Digital-Emerging Grant 101070310).}
    % \thanks{Manuscript received: December, 14, 2022; Revised March, 20, 2022; Accepted May, 8, 2023.}
    % \thanks{Digital Object Identifier (DOI): see top of this page.}
}

% \markboth{IEEE Transactions on Robotics. Preprint Version. Accepted May, 2023}
% {L\"ow \MakeLowercase{\textit{et al.}}: Geometric Algebra for Optimal Control} 

\maketitle

\begin{abstract}
    Many problems in robotics are fundamentally problems of geometry, which lead to an increased research effort in geometric methods for robotics in recent years. The results were algorithms using the various frameworks of screw theory, Lie algebra and dual quaternions. A unification and generalization of these popular formalisms can be found in geometric algebra. The aim of this paper is to showcase the capabilities of geometric algebra when applied to robot manipulation tasks. In particular the modelling of cost functions for optimal control can be done uniformly across different geometric primitives leading to a low symbolic complexity of the resulting expressions and a geometric intuitiveness. We demonstrate the usefulness, simplicity and computational efficiency of geometric algebra in several experiments using a Franka Emika robot. The presented algorithms were implemented in c++20 and resulted in the publicly available library \textit{gafro}. The benchmark shows faster computation of the kinematics than state-of-the-art robotics libraries.
\end{abstract}


% % % % % % % % % % % % % % % % % % % % % % % % % % % % % % % % % % % % % % % % % % % % §

\section{Introduction}
\label{sec:introduction}
    Robot manipulators are used within an increased diversity of environments and tasks, which leads to a large increase in not only the complexity of the surroundings but also in the systems, that need to be able to adapt to different situations. To ensure safe and efficient interaction the corresponding algorithms need to be fast and be based on accurate models of the environment, which makes it important to think carefully about the representations that are used. Many robotics problems are fundamentally problems of geometry, which is why a lot of recent research is focusing on representing and utilizing these geometric properties for solving a wide variety of problems more efficiently. Screw theory, Riemannian geometry, Lie algebra and dual quaternions are just a few examples of the different methods that have been proposed to be used in robotics. Traditionally the kinematics and dynamics of the robots are expressed in different algebras, including linear algebra, vector calculus and quaternion algebra. While quaternions offer a way to avoid the singularities caused by Euler angles, they do not contain position information and thus conversion operations between algebras are required. To address this limitation, dual quaternions were proposed to extend quaternions by a dual unit, resulting in a translation and rotation. We propose in this paper to use geometric algebra instead, which can be seen as a further unification and generalization of these concepts. In particular conformal geometric algebra is a direct extension of dual quaternions \cite{kamarianakisAllInOneGeometricAlgorithm2021}.

    Geometric Algebra (GA) can be seen as a \textit{high-level mathematical language} for geometry that unifies several known concepts, which makes it a very effective tool when the physics of a system need to be modeled. The roots of geometric algebra can be found in Clifford algebra, which was a unification of quaternions and Grassmann algebra \cite{breuilsNewApplicationsClifford2022}. The result was the geometric product, which is the sum of an inner and an outer product. This unfamiliar concept actually leads to algebraic tools that allow for the simplification of many otherwise complex equations, making them more intuitive to handle. A well-known example for this simplification are the Maxwell equations, which reduce to only a single equation in geometric algebra $\left( \bm{\nabla} + \frac{1}{c} \frac{\partial}{\partial t} \right) F = J$ \cite{jootGeometricAlgebraElectrical2019}.

    % % % %
    \begin{figure}[!t]
        % % % %
        \centering
        % % % % 
        \includegraphics[width=\linewidth]{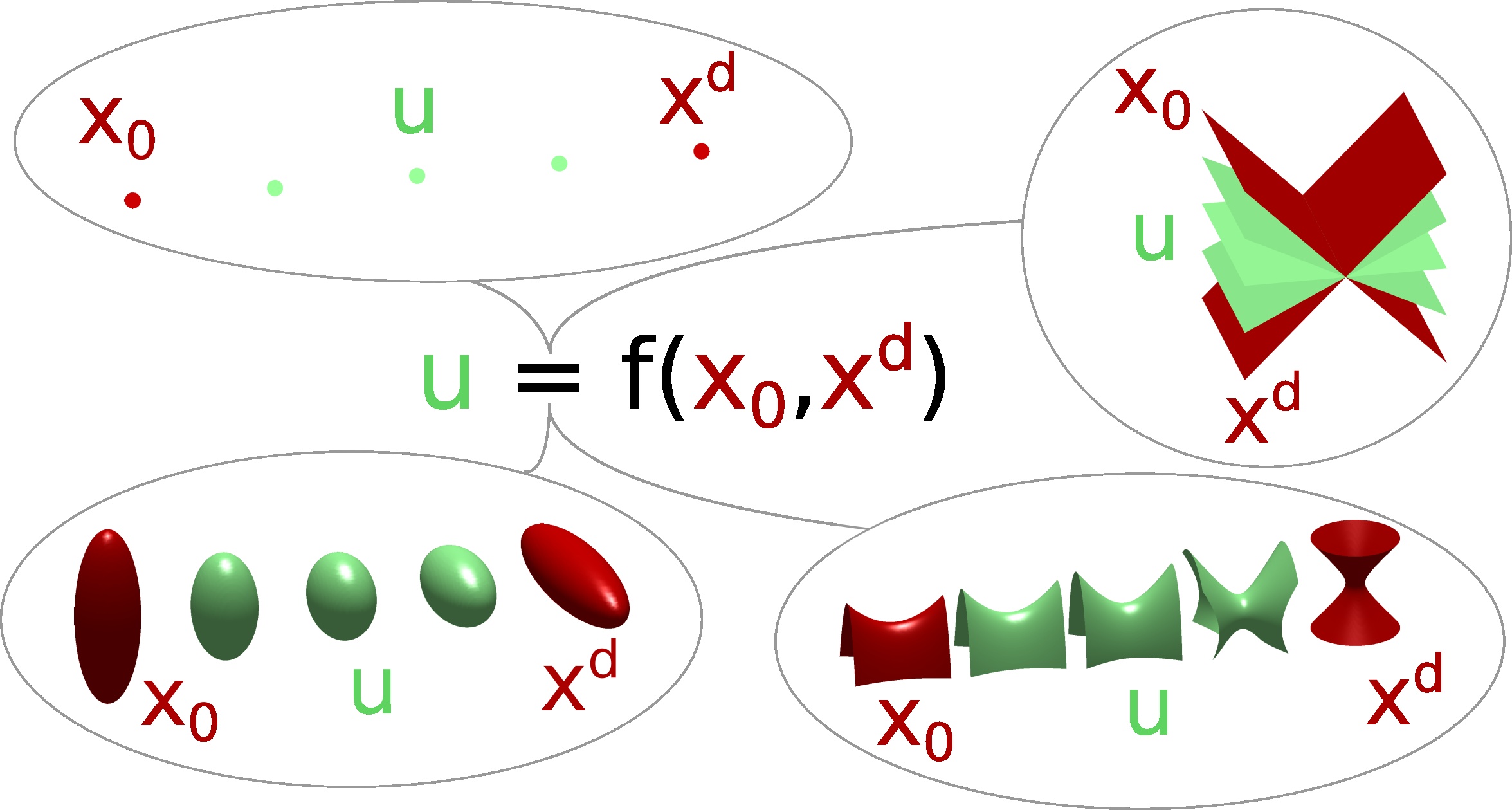}
        % % % % 
        \caption{A generic cost function in geometric algebra can consider different geometric primitives without changing its structure.}
        % % % %  
        \label{fig:generic_cost_function_in_geometric_algebra}
    \end{figure}

    GA is based on a multiplication operation called the geometric product, composed of an inner product and an outer product. The latter describes an oriented plane/volume that extends and generalizes the cross product that is restricted to only 3
    dimensions. The resulting elements are called multivectors. This representation can be used to encode geometric primitives in a uniform manner (depicted in red in Figure \ref{fig:generic_cost_function_in_geometric_algebra}), such as points, lines, planes, spheres, or quadric surfaces such as ellipsoids, as well as the associated transformations $u$ to move from an initial state $x_0$ to a desired state $x_d$, which are called motors (depicted in green in Figure \ref{fig:generic_cost_function_in_geometric_algebra}). In robotics, these operations allow translations and rotations to be treated in the same way, without requiring us to switch between different algebras, as is classically done when handling position data in a Cartesian space and orientation data as quaternions. Practically, GA allows geometric operations to be computed in a very fast way, with compact codes. In Figure \ref{fig:generic_cost_function_in_geometric_algebra} , it means that $u=f(x_0, x_d)$ can be described uniquely for the different geometric objects represented in the figure. 

    The representational advantage of geometric algebra is the geometric significance of its elements, meaning that an object can directly represent geometric primitives, such as lines, spheres and planes, as well as orthogonal transformations, such as rotations, translations, scaling and projections. This allows the direct extraction of geometric information about the problem from the equations. Furthermore, its elements, called multivectors, avoid the parameter redundancy of other representations such as matrices, leading to less memory consumption and optimized computation compared to analytic geometry or vector calculus, which makes it an amenable framework for real-time applications. In engineering the validity of equations is usually determined by a dimensional check of the quantities of the formula. These quantities are of a certain algebraic order when using geometric algebra, which adds a structural check for the validity. These properties were some fundamental criteria in the design of geometric algebra, along with the possibility to formulate basic equations in a coordinate-free manner and to smoothly transfer information between formalisms \cite{hestenesCliffordAlgebraGeometric1985}.

    We want to show the versatility of geometric algebra and that it can be used as a single tool to solve a variety of problems due to its unification of concepts. Simultaneously, we want to make geometric algebra more accessible by enabling the usage of standard tools and solvers for minimizing cost functions that are expressed in geometric algebra. Our contributions are as follows:
    % % % % 
    \begin{itemize}
        \item We extend the Lagrangian dynamics of serial manipulator in conformal geometric algebra to include a non-trivial inertia tensor and subsequently use the derived dynamics in an inverse dynamics control scheme.
        \item We propose the usage of geometric algebra to define objective functions for optimizations as they appear in inverse kinematics and optimal control problems for manipulation tasks and show the modeling of different geometric relations in an optimization problem, while keeping the structure of the cost function uniform. 
        \item We demonstrate how geometric algebra formulations can be seamlessly used within existing frameworks based on linear matrix algebra by exploiting the sparsity of geometric algebra in order to facilitate its adoption. 
        \item We provide an open-source library called \textit{\textbf{g}eometric \textbf{a}lgebra \textbf{f}or \textbf{ro}botics (gafro)} that implements all the presented formulations for robotics. The library is based on a fast and efficient custom implementation of conformal geometric algebra using expression templates. The benchmark shows faster computation of the kinematics than state-of-the-art robotics libraries.
    \end{itemize}

    This paper is organized as follows: Section \ref{sec:related_work} presents the related work, Section \ref{sec:geometric_algebra} introduces geometric algebra in a formal manner, Section \ref{sec:geometric_algebra_for_serial_manipulators} explains how geometric algebra can be used to compute the kinematics and dynamics of serial manipulators, Section \ref{sec:geometric_algebra_for_optimal_control} introduces geometric algebra for optimal control and Section \ref{sec:experiments} then shows the experiments. The source code of the library as well as more information and accompanying videos can be found on our website \newline \texttt{\href{https://tloew.gitlab.io/geometric_algebra/}{tloew.gitlab.io/geometric\_algebra/}}.
% section sec:introduction (end)

% % % % % % % % % % % % % % % % % % % % % % % % % % % % % % % % % % % % % % % % % % % % 

\section{Related Work}
\label{sec:related_work}
    The resurgence of geometric methods in robotics has spawned a variety of different approaches to formalize control, learning and optimization problems in robotics. These methods include screw theory, Riemannian geometry, Lie algebra and dual quaternions. A common motivation between these frameworks is the modeling of robot kinematics and dynamics, which is closely tied to representing rigid body transformations. Geometric algebra essentially presents a generalization and unification of these concepts and thus it is naturally connected to a large variety of recent work in robotics research. 

    % \cite{featherstoneRigidBodyDynamics2008}
    % \cite{mullerScrewLieGroup2018}

    % A common representation is using a Euclidean position vector and quaternion, which essentially corresponds to the manifold $\mathbb{R}^3\times \mathcal{S}^3$. Other methods include the use 4-dimensional matrices representing a rotation and a translation, as is for example the case in screw {}theory \rr. All these methods are essentially representations of the \textit{SE(3)} manifold \rr. 

    % Poses are often represented as a Euclidean position and a quaternion, which is difficult to handle jointly, since it requires the combination of Euclidean with non-Euclidean data. This combination makes it necessary to convert elements between different algebras for the computation, which can result in numerical instabilities.

    Conceptually, geometric algebra can be seen as an extension and generalization of dual quaternions \cite{bayro-corrochanoSurveyQuaternionAlgebra2021}, since dual quaternions can be identified with a certain Clifford algebra, which forms the foundation of geometric algebra. The literature on dual quaternions is hence the most closely related to our work. Starting with formulating rigid body transformations, dual quaternion algebra offers efficient ways for blending them, which is useful in computer graphics \cite{kavanDualQuaternionsRigid2006}. In robotics, there have been various works describing the kinematics and dynamics of robots in dual quaternion algebra such as \cite{leclercq3DKinematicsUsing2013} and \cite{afonsosilvaDynamicsMobileManipulators2022} that combine the geometric understanding of screw theory, the thoroughness of Lie Algebra and the simplicity of spatial algebra. This is also true for the motors of geometric algebra due to the close connection to dual quaternions. Efficient control is an important aspect for using real robots and dual quaternions have been used to design admittance controller \cite{fonsecaCoupledTaskSpaceAdmittance2020} and LQR controller for trajectory tracking \cite{marinhoDualQuaternionLinearquadratic2015}. Collision avoidance is an important aspect of control and in \cite{marinhoDynamicActiveConstraints2019} vector field inequalities based on dual quaternions were proposed to handle them during surgical tasks.

    Along with the regained interest of using geometric methods in robotics came various works proposing the use of differential and Riemannian geometry for learning and optimization problems. The topic of learning from demonstration often requires data to be represented as distributions, thus \cite{calinonGaussiansRiemannianManifolds2020} presented how to use Gaussians on Riemannian manifolds. The manifold of semi-positive definite matrices has been used to study manipulability ellipsoids for learning robot skills \cite{jaquierGeometryawareManipulabilityLearning2021}. In \cite{maricRiemannianMetricGeometryaware2021} a Riemannian metric was proposed that helps manipulators avoid singularities. Riemannian optimization is also used to solve the inverse kinematics problem of kinematic chains using distance geometry \cite{maricRiemannianOptimizationDistanceGeometric2022}. 

    Geometric algebra has been applied successfully in a variety of different applications and fields. For example in the field of computer graphics, which started to re-popularize it, it found applications in mesh deformation \cite{belonApplicationsConformalGeometric2013} as well as ray casting and surface representation \cite{hadfieldExploringNovelSurface2021}. In the domain of image processing techniques for adaptive filtering have been devised \cite{lopesGeometricAlgebraAdaptiveFilters2019}, \cite{heNovelAdaptiveFiltering2020}. 

    A popular example in robotics to show the strengths of geometric algebra is solving the inverse kinematics problem. There have been various methods that proposed to utilize the geometric primitives and their intersection such as FABRIK \cite{aristidouInverseKinematicsSolutions2011} which finds an iterative solution and has also been extended to include model constraints \cite{aristidouExtendingFABRIKModel2016}. Recently another extension, called FABRIKx \cite{kolpashchikovFABRIKxTacklingInverse2022}, was proposed to address the inverse kinematics problem of continuum robots. A similar approach that finds a closed-form solution using the geometric primitive intersection has been described in \cite{zaplanaClosedformSolutionsInverse2022}, while \cite{bayro-corrochanoDifferentialInverseKinematics2007} presented the differential and inverse kinematics of robots using conformal geometric algebra and \textcolor{black}{an iterative inverse kinematics solution was derived in \cite{lechuga-gutierrezIterativeInverseKinematics2022}}.
    Recent work has approached the topic of formulating constrained dynamics in conformal geometric algebra \cite{hadfieldConstrainedDynamicsConformal2020}. \textcolor{black}{Newton-Euler modeling has been proposed for multi-copters using motor algebra in \cite{arellano-muroNewtonEulerModeling2020} and for robot control using conformal geometric algebra in \cite{bayro-corrochanoNewtonEulerModeling2022}.}
    The interpolation of motors, i.e. rigid body transformations, has been shown to have useful applications in surgical robotics to model and plan surgical paths using virtual reality \cite{bayro-corrochanoGeometricIntuitiveTechniques2020}. Conformal geometric algebra was presented for robust pose control of manipulators in \cite{gonzalez-jimenezRobustPoseControl2014} and \cite{zamora-esquivelRobotObjectManipulation2011} used it for robot object manipulation. \textcolor{black}{Building on this work, we develop an optimal control framework that allows the formulation of various geometric primitives as task objectives via a generic cost function. Our extension of the dynamics then enables this optimal control framework to be used in MPC fashion for a torque-controlled robot.}

    Apart from our theoretical contributions we also provide an open-source library that implements all the presented formulations and algorithms. To this end, we have implemented the geometric algebra from scratch using expression templates. There have been various works that published implementations of geometric algebra such as GATL \cite{fernandesExploringLazyEvaluation2021}, GARAMON \cite{breuilsGaramonGeometricAlgebra2019}, Gaigen \cite{fontijneGaigenGeometricAlgebra2006}, TbGAL \cite{sousaTbGALTensorBasedLibrary2020}, GAL \cite{jeremy_ong_2019} and Versor \cite{colapinto2011versor}. These libraries all have in common that they are meant to be generic geometric algebra implementations focusing on the computational and mathematical aspects of the algebra itself. In contrast to that, our implementation is targeted specifically at robotics applications and thus not only implements the low-level algebraic computations but also features the computation of the kinematics and dynamics of serial manipulators as well as generic cost functions for optimal control. We therefore have a similar objective as the DQ robotics \cite{adornoDQRoboticsLibrary2021} library, but using the more general conformal geometric algebra as opposed to dual quaternions.
% section sec:related_work (end)

% % % % % % % % % % % % % % % % % % % % % % % % % % % % % % % % % % % % % % % % % %

\section{Geometric Algebra}
\label{sec:geometric_algebra} 
    In this section, we will give a brief introduction to geometric algebra with a focus on the specific variant known as conformal geometric algebra (CGA). We will use the following notation throughout the paper: $x$ to denote scalars, $\bm{x}$ for vectors, $\bm{X}$ for matrices, $X$ for multivectors and $\bm{\mathcal{X}}$ for matrices of multivectors.
    
    Geometric algebra is a single algebra for geometric reasoning, alleviating the need of utilizing multiple algebras to express geometric relations \cite{bayro-corrochanoGeometricAlgebraApplications2020}. The core idea of geometric algebra is its multiplication operation called the geometric product
    \begin{equation}\label{eq:geometric_product}
        \geometricproduct,
    \end{equation}
    which is the sum of an inner $\cdot$ and an outer $\outer$ product \cite{hestenesCliffordAlgebraGeometric1984}.

    The resulting algebra essentially includes $\mathbb{R}$ and the subspaces of the associated vector space as elements of computations \cite{macdonaldSurveyGeometricAlgebra2017}. Hence, let $\mathbb{R}^{p,q,r}$ be a vector space, where $p,q$ and $r$ are the number of basis vectors that square to 1,-1 and 0, respectively, i.e. the dimension of this vector space is $n=p+q+r$. The associated geometric algebra $\mathbb{G}_{p,q,r}$ then has $2^n=2^{p+q+r}$ basis elements which are called blades. A general element in geometric algebra is called a multivector and is the linear combination of basis blades. This high dimension looks to be leading to an increased complexity, in practice, however, these multivectors usually are very sparse, a fact that we exploit in our implementation. Common variants of geometric include motor algebra $\mathbb{G}^+_{3,0,1}$, projective geometric algebra (PGA) $\mathbb{G}_{3,0,1}$ and conformal geometric algebra (CGA) $\mathbb{G}_{4,1,0}$ \cite{gunnGeometricAlgebrasEuclidean2017}. Dual quaternions can be identified with the Clifford algebra $Cl^+_{0,3,1}$ \cite{johansenDualQuaternionControl2019}, which means they are based on an algebra with a degenerate metric. Due to this, many of the operations that we are presenting in this work are not possible in dual quaternion algebra. 

    In this paper we are using conformal geometric algebra (CGA) \cite{bayro-corrochanoGeometricAlgebraApplications2019}. Conformal refers to angle-preserving transformations. It embeds the 3-dimensional Euclidean space $\mathbb{R}^3$ into the 5-dimensional one $\mathbb{R}^{4,1}$. The corresponding geometric algebra $\cga$ introduces two null-vectors ($\gae{0}$ and $\gae{\infty}$) that essentially represent a point at the origin and a point at infinity. The 5-dimensional space means that CGA has 32 basis blades. An explanation of that structure can be found in Appendix \ref{sub:structure_of_a_multivector_in_conformal_geometric_algebra}.
    
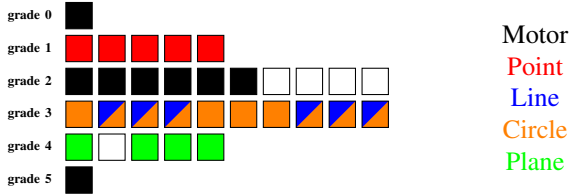
\begin{figure}[!ht]
        % % % %
        \centering
        % % % % 
        \begin{minipage}{0.55\linewidth}
            \centering
            % % % % 
            \begin{tikzpicture}[scale=1.75]
                \draw[draw=white] (-1.5*0.25,0*0.25) rectangle ++(0.2,0.2) node [pos=0.5] {\tiny\textbf{grade 0}};
                \draw[draw=white] (-1.5*0.25,-1*0.25) rectangle ++(0.2,0.2) node [pos=0.5] {\tiny\textbf{grade 1}};
                \draw[draw=white] (-1.5*0.25,-2*0.25) rectangle ++(0.2,0.2) node [pos=0.5] {\tiny\textbf{grade 2}};
                \draw[draw=white] (-1.5*0.25,-3*0.25) rectangle ++(0.2,0.2) node [pos=0.5] {\tiny\textbf{grade 3}};
                \draw[draw=white] (-1.5*0.25,-4*0.25) rectangle ++(0.2,0.2) node [pos=0.5] {\tiny\textbf{grade 4}};
                \draw[draw=white] (-1.5*0.25,-5*0.25) rectangle ++(0.2,0.2) node [pos=0.5] {\tiny\textbf{grade 5}};

                \draw[draw=black,fill=black] (0,0*0.25) rectangle ++(0.2,0.2);

                \draw[draw=black,fill=red] (0*0.25,-1*0.25) rectangle ++(0.2,0.2);
                \draw[draw=black,fill=red] (1*0.25,-1*0.25) rectangle ++(0.2,0.2);
                \draw[draw=black,fill=red] (2*0.25,-1*0.25) rectangle ++(0.2,0.2);
                \draw[draw=black,fill=red] (3*0.25,-1*0.25) rectangle ++(0.2,0.2);
                \draw[draw=black,fill=red] (4*0.25,-1*0.25) rectangle ++(0.2,0.2);
                
                \draw[draw=black,fill=black] (0*0.25,-2*0.25) rectangle ++(0.2,0.2);
                \draw[draw=black,fill=black] (1*0.25,-2*0.25) rectangle ++(0.2,0.2);
                \draw[draw=black,fill=black] (2*0.25,-2*0.25) rectangle ++(0.2,0.2);
                \draw[draw=black,fill=black] (3*0.25,-2*0.25) rectangle ++(0.2,0.2);
                \draw[draw=black,fill=black] (4*0.25,-2*0.25) rectangle ++(0.2,0.2);
                \draw[draw=black,fill=black] (5*0.25,-2*0.25) rectangle ++(0.2,0.2);
                \draw[draw=black] (6*0.25,-2*0.25) rectangle ++(0.2,0.2);
                \draw[draw=black] (7*0.25,-2*0.25) rectangle ++(0.2,0.2);
                \draw[draw=black] (8*0.25,-2*0.25) rectangle ++(0.2,0.2);
                \draw[draw=black] (9*0.25,-2*0.25) rectangle ++(0.2,0.2);
                
                \draw[draw=black,fill=orange] (0*0.25,-3*0.25) rectangle ++(0.2,0.2);
                \draw[draw=black,diagonal fill={orange}{blue}] (1*0.25,-3*0.25) rectangle ++(0.2,0.2);
                \draw[draw=black,diagonal fill={orange}{blue}] (2*0.25,-3*0.25) rectangle ++(0.2,0.2);
                \draw[draw=black,diagonal fill={orange}{blue}] (3*0.25,-3*0.25) rectangle ++(0.2,0.2);
                \draw[draw=black,fill=orange] (4*0.25,-3*0.25) rectangle ++(0.2,0.2);
                \draw[draw=black,fill=orange] (5*0.25,-3*0.25) rectangle ++(0.2,0.2);
                \draw[draw=black,fill=orange] (6*0.25,-3*0.25) rectangle ++(0.2,0.2);
                \draw[draw=black,diagonal fill={orange}{blue}] (7*0.25,-3*0.25) rectangle ++(0.2,0.2);
                \draw[draw=black,diagonal fill={orange}{blue}] (8*0.25,-3*0.25) rectangle ++(0.2,0.2);
                \draw[draw=black,diagonal fill={orange}{blue}] (9*0.25,-3*0.25) rectangle ++(0.2,0.2);
                
                \draw[draw=black,fill=green] (0*0.25,-4*0.25) rectangle ++(0.2,0.2);
                \draw[draw=black] (1*0.25,-4*0.25) rectangle ++(0.2,0.2);
                \draw[draw=black,fill=green] (2*0.25,-4*0.25) rectangle ++(0.2,0.2);
                \draw[draw=black,fill=green] (3*0.25,-4*0.25) rectangle ++(0.2,0.2);
                \draw[draw=black,fill=green] (4*0.25,-4*0.25) rectangle ++(0.2,0.2);
                
                \draw[draw=black,fill=black] (0*0.25,-5*0.25) rectangle ++(0.2,0.2);
            \end{tikzpicture}
        \end{minipage}
        % % % % 
        \hfill
        % % % %
        \begin{minipage}{0.39\linewidth} 
            \centering
            \textcolor{black}{Motor}
            \\ \textcolor{red}{Point}
            \\ \textcolor{blue}{Line}
            \\ \textcolor{orange}{Circle}
            \\ \textcolor{green}{Plane}
        \end{minipage}
        % % % % 
        \caption{\textcolor{black}{Non-zero elements of various geometric primitives in their primal representations in conformal geometric algebra. Boxes represent basis blades and colored boxes represent the non-zero blades of the geometric primitive with the matching color. It can be seen that of the 32 basis blades only a sparse number is used for the representations. Note that geometric primitives are single-grade objects, while transformations are mixed-grade. The boxes correspond to the basis blades that are shown in Table \ref{tab:basis_blades_of_conformal_geometric_algebra} in Appendix \ref{sub:structure_of_a_multivector_in_conformal_geometric_algebra}.}}
        % % % % 
        \label{fig:structure_of_various_geometric_primitives_in_conformal_geometric_algebra_}
    \end{figure}

    A point $\bm{x}$ in Euclidean space $\mathbb{R}^3$ is embedded into CGA by using the conformal embedding, which is bijective, meaning that any point $\bm{x}\in \mathbb{R}^3$ can be uniquely identified with a point $X\in\cga$
    \begin{equation}\label{eq:conformal_embedding}
        X = \conformalembedding.
    \end{equation}
    Points are the basic geometric primitives that can be used to construct others by the spanning operation of the outer product. These geometric primitives are in general nullspace representations with respect to either the inner (IPNS) or the outer (OPNS) product, meaning that a geometric primitive is defined by the set of all Euclidean points that result in zero upon multiplication when embedded in CGA, i.e. 
    \begin{eqnarray}\label{eq:product_nullspaces}
        &\innerproductnullspace,
        \\
        &\outerproductnullspace.
    \end{eqnarray}
    The IPNS and OPNS representations are connected by a duality relationship. Duality in this case means multiplication with the pseudoscalar $I$, the highest grade element of the algebra (i.e. $I=e_{0123\infty}$ for CGA)
    \begin{equation}\label{eq:duality}
        X^* = I X.
    \end{equation}
    We refer to the OPNS as the primal space for its more convenient usage, which consequently makes the IPNS the dual representation, although both representations can be used to represent all geometric primitives. \textcolor{black}{In Figure \ref{fig:structure_of_various_geometric_primitives_in_conformal_geometric_algebra_}, we show the subspaces that several geometric primitives occupy within the algebra to demonstrate their sparsity. The type of a multivector resulting from a product operation can thus be determined by looking at the expected non-zero elements of the expression, which is exploited in our implementation of GA.} The primitives of conformal geometric algebra with their corresponding equations for construction can be found in Appendix \ref{sec:geometric_primitives_in_conformal_geometric_algebra}. These primitives can be extended when going to higher dimensional geometric algebras, for example, $\mathbb{G}_{6,3}$ additionally introduces quadric surfaces like ellipsoids and hyperboloids \cite{zamora-esquivelGeometricAlgebraDescription2014} and the Quadric Conformal Geometric Algebra (QCGA) $\mathbb{G}_{9,3}$ allows using arbitrary quadric surfaces \cite{breuilsQuadricConformalGeometric}. 

    % % % %

    Conformal geometric algebra provides an exception-free way of computing incidence relations between geometric objects \cite{bayro-corrochanoComputingConformalSpace2022}. This is achieved via the meet operator
    \begin{equation}\label{eq:meet_operator}
        \gameet.
    \end{equation}
    The resulting multivector retains a geometric meaning, e.g. when a line meets a sphere there are three possibilities: 
    % % % % 
    \begin{itemize}
        \item the line intersects the sphere, in which case $Y$ is a point pair;
        \item the line is tangential to the sphere, which results in a single point;
        \item the line and the sphere are completely separate, resulting in an imaginary point that is related to the distance between the objects.
    \end{itemize}
    This geometric result is directly encoded in the result and no special cases need to be considered.

    There are several geometric operations available in geometric algebra such as translations and rotations, but also dilations, reflections, projections and rejections. For brevity we only present rigid body transformations, i.e. translations and rotations, in this paper. Thorough introductions can be found in \cite{jootGeometricAlgebraElectrical2019} and \cite{perwassGeometricAlgebraApplications2009} and a survey of relevant research in \cite{bayro-corrochanoSurveyQuaternionAlgebra2021} and \cite{hitzerCurrentSurveyClifford2022}.

    % % % % % % % % % % % % % % % % % % % % % % % % % % % % % % % % % % % % % % % % % %
    
    \subsection{Rigid Body Motions in Geometric Algebra}
    \label{sub:rigid_body_motions_in_geometric_algebra}
        The elements that describe rigid body motions are called rotors, translators and more generally: motors. A general motor is hence composed of a translator and a rotor (we omit other conformal operations such as scaling in this introduction to geometric algebra in order to keep it short), i.e. 
        \begin{equation}\label{eq:general_motor}
            M = TR.
        \end{equation}
        A motor applied to multivectors results in a sandwiching product, similar to how quaternions rotate vectors        
        \begin{equation}
            Y = MX\reverse{M},
        \end{equation}
        \textcolor{black}{where $\reverse{M}$ stands for the reverse of a motor, which can thought of as being similar to a conjugate quaternion\footnote{\textcolor{black}{Note the similarities with the operations using conjugate quaternion in quaternion algebra.}}.}

        Both translators and rotors can be found with an exponential mapping of bivectors, i.e.
        \begin{equation}\label{eq:cgatranslator}
            \cgatranslator,
        \end{equation}       
        and 
        \begin{equation}\label{eq:cgarotor}
            \cgarotor.
        \end{equation}
        The bivectors are $\bm{t}\outer\ei\in\text{span}\left\{\bm{e}_{1\infty},\bm{e}_{2\infty},\bm{e}_{3\infty}\right\}$ and $B\in\text{span}\left\{\bm{e}_{23},\bm{e}_{13},\bm{e}_{13}\right\}$, respectively. Rotors can be seen as isomorphic to quaternions, they are however more general and do not require the introduction of complex numbers.

        The motors in geometric algebra form a group, which is an even sub-algebra $\mathbb{M}$ of $\cga^+$
        \begin{equation}\label{eq:motorspace}
            \motorspace.
        \end{equation}
        It forms a Lie group with an associated Lie algebra, which is the bivector algebra in the linear subspace $\mathbb{B}$ that is defined as 
        \begin{equation}\label{eq:bivectorsubspace}
            \bivectorsubspace.
        \end{equation}
        Since the motor group is a Lie group it is also a smooth manifold. The motor manifold $\mathcal{M}$ can be found with the group constraint
        \begin{equation}\label{eq:motormanifold}
            \motormanifold.
        \end{equation}
        Motors are isomorphic to dual quaternions \cite{perwassGeometricAlgebraApplications2009}, which makes them also isomorphic to $SE(3)$. They represent, however, a more general concept of transformations that is valid in any dimension. Furthermore, due to their similarities with dual quaternions the same advantages over transformation matrices apply to motors as well, i.e. they require less memory and operations for multiplication compared to transformation matrices \cite{adornoTwoarmManipulationManipulators2011}. The motor manifold being a Lie group consequently makes the bivector algebra its Lie algebra. The operation that connects the motor manifold with the bivector algebra is the exponential map, and its inverse the logarithmic map
        \begin{equation}\label{eq:motor_exp_log_map}
            M = \exp(B)
            \hspace{3mm}
            \iff
            \hspace{3mm}
            B = \log(M).
        \end{equation}
        \textcolor{black}{The exponential and logarithmic maps are the Lie group operators that move an element from the bivector Lie algebra $\mathbb{B}$ (equivalent to $\mathfrak{se}(3)$) to the motor manifold Lie group $\mathcal{M}$ (equivalent to $SE(3)$) and vice versa.} The interpretation of the bivector is the representation of a dual line $B = L^*$. This essentially defines the screw axis of the motor. Note that $L^*$ is the IPNS of a line and as such a 2-blade, i.e. a bivector \cite{tingelstadAutomaticDifferentiationOptimization2017}. The motor manifold therefore elegantly combines position and orientation. Having both position and orientation represented by one entity removes the need to switch between algebras for computation, e.g. linear and quaternion algebra. Using the motor manifold optimization problems can be solved as presented in \cite{tingelstadMotorParameterization2018}.

\begin{figure}[!ht]
            % % % %
            \centering
            % % % % 
            \begin{subfigure}[]{0.3\linewidth}
                \centering
                \includegraphics[width=\linewidth]{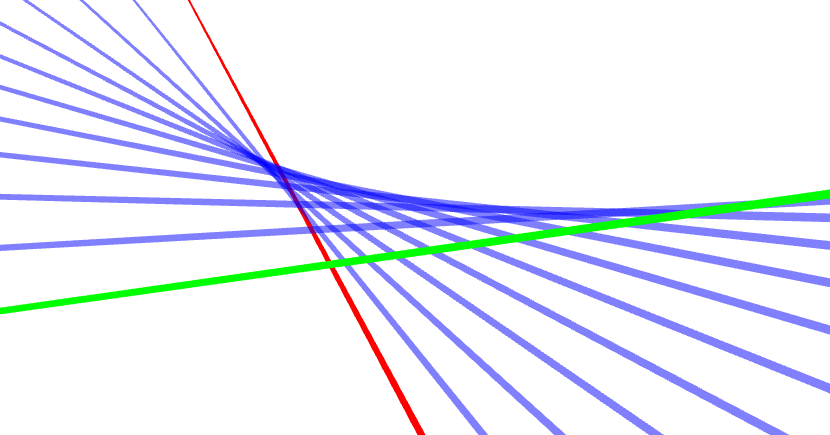}
            \end{subfigure}
            \begin{subfigure}[]{0.3\linewidth}
                \centering
                \includegraphics[width=\linewidth]{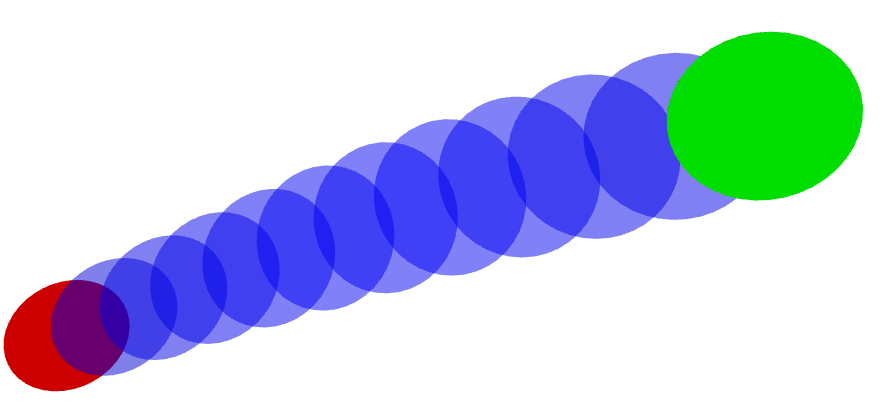}
            \end{subfigure}
            \begin{subfigure}[]{0.3\linewidth}
                \centering
                \includegraphics[width=\linewidth]{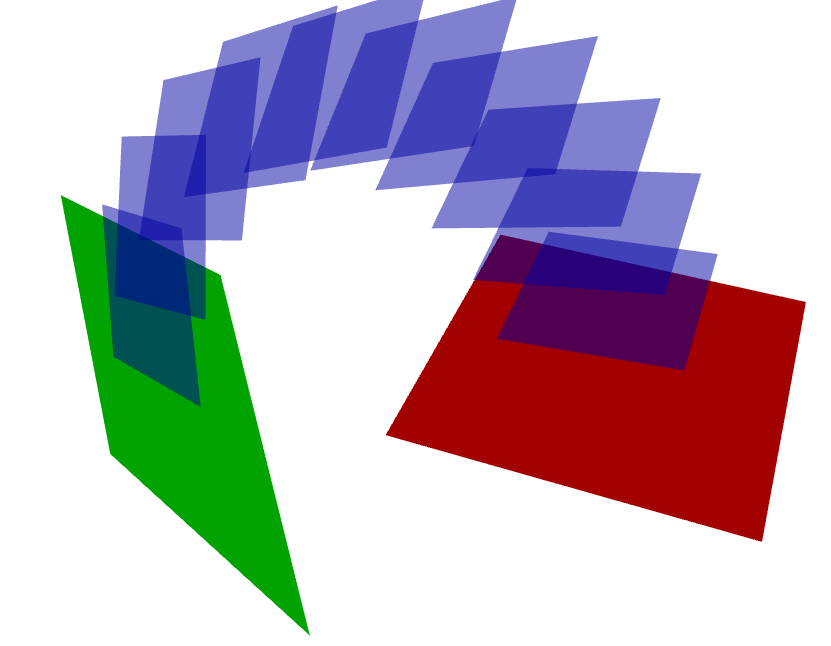}
            \end{subfigure}
            % % % % 
            \caption{\textcolor{black}{Rigid body transformations of various geometric primitives. Red marks the initial primitive and green the final one, the primitives resulting from the trajectory of interpolated motors are shown in blue.}}
            % % % % 
            \label{fig:rigid_body_transformations_of_various_geometric_primitives_}
        \end{figure}

        Since the motors and the geometric primitives are part of the same algebra, the motors can be used to apply rigid body transformations to these primitives. This means that the primitives can be translated, rotated, reflected and scaled using angle-preserving transformations. \textcolor{black}{Some visual examples of how motors are transforming geometric primitives are shown in Figure \ref{fig:rigid_body_transformations_of_various_geometric_primitives_}.}
        % % % %

        It has been shown in \cite{bayro-corrochanoGeometricIntuitiveTechniques2020} that efficient interpolation between motors can be achieved via the bivector space, which is similar to spherical linear interpolation (SLERP). In this case the parameterized motor curve $M(t)$ can be found as an interpolation in the bivector space of viapoint motors using the exponential and logarithmic map
        \begin{equation}\label{eq:motorcurve}
            \motorcurve.
        \end{equation}
        The weights need to fulfill $\sum_{j=1}^{n} w_j = 1$ for each timestep, but can otherwise be chosen arbitrarily. In \cite{belonApplicationsConformalGeometric2013} this is exploited for mesh deformation.
% section geometric_algebra (end)

\section{Geometric Algebra for Serial Manipulators}
\label{sec:geometric_algebra_for_serial_manipulators}

    In this section, we present how geometric algebra can be used to express the kinematics and dynamics of serial manipulators. We do this while also explicitly drawing connections to the expressions and terminology of classical linear algebra.   

    % % % % % % % % % % % % % % % % % % % % % % % % % % % % % % % % % % % % % % % %
    
    \subsection{Forward Kinematics of Serial Manipulators}
    \label{sub:forward_kinematics_of_serial_manipulators}
        The forward kinematics of a kinematic chain of $N$ joints can easily be defined using motors \textcolor{black}{\cite{bayro-corrochanoMotorAlgebraApproach2000}}. Assuming that we only have revolute joints, the forward motor $M (\posjoint)$, given the configuration $\posjoint$, can be computed with
        \begin{equation}\label{eq:cgaforwardkinematics}
            M (\posjoint) = \prod_{i=1}^{N} M_i(q_i) = \prod_{i=1}^{N} M_{F,i}R_i(q_i).
        \end{equation}
        The constant joint-specific motors $M_{F,i}$ represent the local frames of the joints with the rotation in that frame expressed by the rotor
        \begin{equation}
            R_i(q_i) = \exp \left(-\frac{1}{2}q_i B_i\right),
        \end{equation}
        where the bivectors $B_i$ essentially represent the rotation planes of the joints. These quantities can easily be found using e.g. DH-parameters \cite{sicilianoRobotics2009}.

        We denote by $M^k(\posjoint)$ the forward kinematics up the $k$-th joint, i.e.
        %source
        \begin{equation}\label{eq:cgaforwardkinematicstok}
            M^k(\posjoint) = \prod_{i=1}^{k} M_i(q_i) = \prod_{i=1}^{k} M_{F,i}R_i(q_i).
        \end{equation}
        
    % subsection ssub:robot_forward_kinematics_using_dh_parameters (end)

    % % % % % % % % % % % % % % % % % % % % % % % % % % % % % % % % % % % % % % % % % %
    
    \subsection{Jacobians of Serial Manipulators}
    \label{sub:jacobians_of_serial_manipulators}
        In the literature about serial kinematic chains one can generally find the distinction between two Jacobians: the geometric and the analytic Jacobian. In this section, we will explain how these quantities translate to geometric algebra.

        Using an arbitrary representation of the end-effector forward kinematics 
        \begin{equation}\label{eq:arbitrary_forward_kinematics}
            \bm{\xi} = \bm{f}(\posjoint),
        \end{equation}
        the analytic Jacobian is defined as the partial derivatives of the forward kinematic function $\bm{f}(\posjoint)$ w.r.t. the joint angles, i.e.
        \begin{equation}\label{eq:definition_analytic_jacobian}
            \ajacobian = \frac{\partial \bm{f}(\posjoint)}{\partial \posjoint},
        \end{equation}
        and it relates the joint angle velocity to time-derivatives of the end-effector configuration using the given representation
        \begin{equation}\label{eq:analytic_jacobian_velocity_relationship}
            \dot{\bm{\xi}} = \ajacobian \veljoint.
        \end{equation}
        The geometric Jacobian on the other hand defines the relationship of the joint angle velocity to the linear and angular velocity of the end-effector in a certain coordinate frame
        \begin{equation}\label{eq:geometric_jacobian_velocity_relationship}
            \begin{bmatrix}\bm{v} \\ \bm{\omega}\end{bmatrix}  = \gjacobian \veljoint.
        \end{equation}
        The relationship between the analytic and the geometric Jacobians can be found by a representation specific mapping 
        \begin{equation}\label{eq:relationship_analytic_geometric_jacobian}
            \gjacobian = \bm{J}^M(\bm{\xi}) \ajacobian.
        \end{equation}

        In geometric algebra the analytic Jacobian can be found as the derivative of the forward kinematics motor defined in Equation \eqref{eq:cgaforwardkinematics}, i.e.
        \begin{equation}\label{eq:ga_analytic_jacobian}
            \gaajacobian = \frac{\partial M(\posjoint)}{\partial \posjoint} = \mat{\frac{\partial M(\posjoint)}{\partial q_1} \ldots \frac{\partial M(\posjoint)}{\partial q_N}}.
        \end{equation}
        \textcolor{black}{Note that the size of the multivector matrix is $\gaajacobian\in \mathbb{M}^{1\times N}\subset \cga^{1\times N}$ with its elements corresponding to motors.} The partial derivative of the forward motor w.r.t the $i$-th joint angle is
        \begin{equation}\label{eq:partial_derivative_forward_motor}
            \frac{\partial M(\posjoint)}{\partial q_i} = M_1(q_1)\ldots M_{F,i} \left(-\frac{1}{2} B_i\right) R_i(q_i) \ldots M_N(q_N).
        \end{equation}

        Similarly, the geometric Jacobian of a serial kinematic chain in geometric algebra can be found by transforming the rotation bivectors of each joint using the respective motor, i.e. 
        \begin{equation}\label{eq:ga_geometric_jacobian}
            \gamatrix{J}^{G}_j (\posjoint) = \begin{bmatrix} B_1^\prime & \ldots & B_j^\prime & \bm{0} \end{bmatrix},
        \end{equation}
        with the rotation bivectors
        \begin{equation}\label{eq:ith_rotation_bivector}
            B_i^\prime = M_1\ldots M_{i-1}M_{F,i}B_i \reverse{M}_{F,i}\reverse{M}_{i-1}\ldots\reverse{M}_1.
        \end{equation}
        \textcolor{black}{In this case the size of the multivector matrix is $\gagjacobian\in \mathbb{B}^{1\times N}\subset \cga^{1\times N}$. Its elements correspond to bivectors.}
        
        From Equations \eqref{eq:partial_derivative_forward_motor} and \eqref{eq:ith_rotation_bivector} the relationship between the analytic and geometric Jacobians in geometric algebra can easily be derived as
        \begin{equation}\label{eq:ga_relationship_analytic_geometric_jacobian}
            \gamatrix{J}^{G}_{kj} (\posjoint) = -2 \gamatrix{J}^{A}_{kj}(\posjoint) \reverse{M}(\posjoint).
        \end{equation}

        For the computation of the dynamics, the time derivative of the geometric Jacobian is also required and it can be found as
        \begin{equation}\label{eq:ga_geometric_jacobian_dt}
            \gadtgjacobian = 
            \mat{
                \dot{B}_1' & 0 & 0 & \cdots & 0
                \\
                \vdots & \dot{B}_2' & 0 & \cdots & 0
                \\
                \vdots & \vdots & \vdots & \ddots & \vdots
                \\
                \dot{B}_1' & \dot{B}_2' & \dot{B}_3' & \cdots & \dot{B}_N'
            }.
        \end{equation}
        The required time derivatives of the rotation bivectors $\dot{B}_j'$ can be found using 
        \begin{equation}\label{eq:ga_rotation_bivectors_dt}
            \gamatrix{\dot{B}}'(\posjoint,\veljoint) = \mat{\dot{B}_1'\\\vdots \\\dot{B}_N'} = \gamatrix{J}^{\times}(\posjoint)\veljoint,
        \end{equation}
        with
        \begin{equation}\label{eq:ga_rotation_bivectors_dt_element}
            \gamatrix{J}_{ij}^{\times}(\posjoint) = B_i' \times B_j' = \frac{1}{2} (B_i' B_j' - B_j' B_i'), 
        \end{equation}
        where the operator $\times$ is called the commutator product. 
        
    % subsection jacobians_of_serial_manipulators (end)

   % % % % % % % % % % % % % % % % % % % % % % % % % % % % % % % % % % % % % % % %
    
    \subsection{Inverse Kinematics of Serial Manipulators}
    \label{sub:inverse_kinematics_of_serial_manipulators}

        Using the expressions derived in Section \ref{sub:jacobians_of_serial_manipulators}, the inverse kinematics problem for a serial kinematic chain can be formulated as an optimization problem on the motor manifold. The goal is to find the joint angles $\posjoint$ that minimize the following equation
        \begin{equation}\label{eq:motor_ik_manifold}
            \posjoint^* = \arg\min_{\posjoint} \Big\|\log\left(\reverse{M}_{\text{target}} M(\posjoint)\right)\Big\|_2^2.
        \end{equation}
        The forward kinematics motor $M(\posjoint)$ can be found using Equation \eqref{eq:cgaforwardkinematics}. The expression $\reverse{M}_2M_1$ can be understood as the shortest screw motion between two points on the motor manifold. The $\log(\cdot)$ operation moves the problem to Bivector space, i.e. the Lie algebra of the motor manifold.

        The Jacobian can be found as
        \begin{equation}\label{eq:ik_partial_derivative}
            \bm{J}_{\mathbb{B}}(\bm{q}) = \mathcal{E}^{\mathbb{B}\to \mathbb{R}^6}\left[\frac{\partial}{\partial q_i} \log\left(\reverse{M}_{\text{target}} M(\posjoint)\right)\right].
        \end{equation}
        Here $\bm{J}_{\mathbb{B}}(\bm{q}) \in \mathbb{R}^{6\times N}$ is a linear algebra matrix. The interpretation of $\bm{J}_{\mathbb{B}}(\bm{q})$ is an embedding of the multivectors into a matrix algebra and exploiting their sparsity. The expression can be further untangled into 
        \begin{equation}\label{eq:jacobian_local}
            \bm{J}_{\mathbb{B}}(\posjoint) = \bm{J}_{\mathcal{M}\to\mathbb{B}}(\posjoint) \bm{J}_{\mathcal{M}}(\posjoint).
        \end{equation}
        $\bm{J}_{\mathcal{M}}(\posjoint) \in \mathbb{R}^{8\times N}$ is the embedding of the analytic Jacobian of Equation \eqref{eq:ga_analytic_jacobian} multiplied by the target motor, i.e. 
        \begin{equation}\label{eq:ga_ik_motor_jacobian}
            \bm{J}_{\mathcal{M}}(\posjoint) = \mathcal{E}^{\mathcal{M}\to \mathbb{R}^8}\left[\reverse{M}_{\text{target}}\gaajacobian\right].
        \end{equation}
        $\bm{J}_{\mathcal{M}\to\mathbb{B}}(\posjoint) \in \mathbb{R}^{6\times 8}$ can be understood as the Jacobian of the local parameterization from the motor manifold to the bivector space and hence is the Jacobian of the $\log(\cdot)$ operation. The derivation of $\bm{J}_{\mathcal{M}\to\mathbb{B}}$ can be found in Appendix \ref{sub:derivation_of_the_jacobian_of_the_motor_logarithmic_map}. 
        
        With Equation \eqref{eq:ik_partial_derivative} the Gauss-Newton step becomes
        \begin{equation}\label{eq:gauss_newton_ik}
            \posjoint_{k+1} = \posjoint_{k} - \alpha \left( \bm{J}_{\mathbb{B}}(\posjoint)^\trsp \bm{J}_{\mathbb{B}}(\posjoint) \right)^{-1} \bm{J}_{\mathbb{B}}(\posjoint)^\trsp \bm{f}(\posjoint_k),
        \end{equation}
        where $\alpha$ is the line-search parameter. This shows how geometric algebra functions on the motor manifold can be optimized using classical methods by embedding the multivectors into a matrix algebra, which will be exploited later when defining the cost functions for optimal control problems in geometric algebra.

    % subsection ssub:inverse_kinematics_on_the_motor_manifold (end)

    % % % % % % % % % % % % % % % % % % % % % % % % % % % % % % % % % % % % % % % % % %
    
    \subsection{Dynamics of Serial Manipulators}
    \label{sub:dynamics_of_serial_manipulators}
        In classical linear algebra, the manipulator equation is found to be
        \begin{equation}\label{eq:la_manipulator_equation}
            \bm{M}(\posjoint)\accjoint + \bm{C}(\posjoint,\veljoint)\veljoint + \bm{g}(\posjoint) = \bm{\tau} - \bm{\tau}_{\text{ext}},
        \end{equation}
        where $\bm{M}(\posjoint)$ is known as the inertia or generalized mass matrix, $\bm{C}(\posjoint,\veljoint)$ is representing Coriolis/centrifugal forces, $\bm{g}(\posjoint)$ stands for the gravitational forces, $\bm{\tau}$ is the vector of joint torques and $\bm{\tau}_{ext}$ are the external torques. 

        %
        % \begin{equation}\label{eq:ga_manipulator_equation}
        %     \gamatrix{M}(\posjoint)\accjoint + \gamatrix{C}(\posjoint,\veljoint)\veljoint + \gamatrix{G}(\posjoint)= \bm{\tau} - \bm{\tau}_{ext}.
        % \end{equation}
        %
        In geometric algebra, Equation \eqref{eq:la_manipulator_equation} can be transformed to a simplified version, which was shown in \cite{bayro-corrochanoGeometricAlgebraApplications2020}. The influence of the link inertia in that work, however, was assumed to be a scalar constant, which of course is not accurate for real systems. Therefore we extend this equation by a joint position dependent inertia tensor and subsequently derive the necessary influence on the Coriolis/centrifugal forces. 
        \textcolor{black}{We first present the geometric algebra reformulation of Equation \eqref{eq:la_manipulator_equation} that was derived in \cite{bayro-corrochanoGeometricAlgebraApplications2020} and then present our extension with the joint position dependent inertia tensor.} The elements of Equation \eqref{eq:la_manipulator_equation} can be expressed as multivector matrices, where the generalized mass matrix becomes 
        \begin{equation}\label{eq:ga_generalized_mass_matrix}
            \gamatrix{M}(\posjoint) = \gamatrix{I}(\posjoint) + \gamatrix{V}^\trsp(\posjoint)\bm{m}\gamatrix{V}(\posjoint).
        \end{equation}
        The scalar valued $\bm{m}$ is an $N\times N$ matrix that contains all link masses along its diagonal.
        The Coriolis/centrifugal forces become
        \begin{equation}\label{eq:ga_corolios_force§}
            \gamatrix{C}(\posjoint,\veljoint)\veljoint = \gamatrix{\dot{I}}(\posjoint,\veljoint) + \gamatrix{V}^\trsp(\posjoint)\bm{m} \gamatrix{\dot{V}}(\posjoint,\veljoint)\veljoint,
        \end{equation}
        and the gravitational forces are
        \begin{equation}\label{eq:ga_graviatational_forces}
            \gamatrix{G}(\posjoint) = \gamatrix{V}^\trsp(\posjoint)\bm{m}\gamatrix{G}.
        \end{equation}
        The constant matrix $\gamatrix{G}\in\cga^{N\times 1}$ contains the gravitational acceleration with the information about the direction. In the usual case, all elements therefore are equal to $g\gae{3}$. The recurring multivector matrix $\gamatrix{V}(\posjoint)\in\cga^{N\times N}$ can be found using the current centers of mass of the links $X_j^{CoM}$ and the current axes of rotation of the joints, expressed as bivectors $B_k'$. An element of this matrix therefore becomes 
        \begin{equation}\label{eq:ga_v_matrix}
            \gamatrix{V}_{j,k} (\posjoint) = X_j^{CoM}(\posjoint)\inner B_k'.
        \end{equation}
        The interpretation of this matrix is the computation of the lever arms of the centers of mass of the links w.r.t. each joint. Its time derivative can be found to be 
        \begin{equation}\label{eq:ga_v_matrix_dt}
            \gamatrix{\dot{V}}_{j,k} (\posjoint,\veljoint) = \Big(\bm{I}\gamatrix{V}(\posjoint)\veljoint\Big) \gagjacobian + \left(\bm{I}\gamatrix{X}^{CoM}(\posjoint)\right) \gadtgjacobian.
        \end{equation}
        Note that $\bm{I}$ in this case is an $N\times N$ identity matrix, such that the expression $\gamatrix{V}(\posjoint)\veljoint$ becomes a square matrix instead of a vector. The same applies to the expression $\bm{I}\gamatrix{X}^{CoM}(\posjoint)$, where the matrix $\gamatrix{X}^{CoM}(\posjoint)\in\cga^{N\times 1}$ contains all centers of masses of the links.
        
        As mentioned, we are not assuming the inertia to be constant in this work, since we want to use the inverse dynamics control scheme in our experiments, which requires an exact computation. Therefore, we have derived the influence of the link inertia, given the current joint state $\gamatrix{I}(\posjoint)$, as well as its time derivative $\dot{\gamatrix{I}}(\posjoint,\veljoint)$. The inertia matrix can be found as a summation over the joint angles of the manipulator, accounting for the influence of each joint, i.e.
        \begin{equation}\label{eq:ga_inertia_matrix}
            \gamatrix{I}(\posjoint) = \sum_{i=0}^N \gamatrix{B}_i^\trsp \mathcal{I}(\gamatrix{B}_i).
        \end{equation}
        The bivector matrix $\gamatrix{B}_{i}$ is a $1\times N$ row-vector and contains the rotation generators of each joint w.r.t the current joint. The $j$-th element of $\gamatrix{B}_i$ can thus be found as
        \begin{equation}\label{eq:ga_inertia_bivector}
            B_{i,j} = \reverse{R}_i(\posjoint) \log^R\Big(\gagjacobianel{ij}\Big) R_i(\posjoint).
        \end{equation}
        The expression $\log^R(\cdot)$ in these equations stands for the logarithmic map of the rotor part of the motor that is the $ij$-th element of the geometric Jacobian. It hence returns a bivector with non-zero elements corresponding to the basis blades $\gae{23},\gae{13}$ and $\gae{12}$. 

        The time derivative of the inertia matrix thus follows as
        \begin{equation}\label{eq:ga_inertia_matrix_dt}
            \gamatrix{\dot{I}}(\posjoint,\veljoint) = \sum_{i=0}^N \gamatrix{B}_i^\trsp \Big{(}\mathcal{I}(\gamatrix{\dot{B}}_i) + \reverse{R}_i(\posjoint) \hat{B}_i^w R_i(\posjoint)\Big{)}.
        \end{equation}
        The time derivative of the rotation generators $\gamatrix{\dot{B}}_{i}$ can be found in the same way as the elements of $\gamatrix{B}_{i}$, but using the time derivative of the geometric Jacobian
        \begin{equation}\label{eq:ga_inertia_bivector_dt}
            \dot{B}_{i,j} = \reverse{R}_i(\posjoint) \log^R\Big(\gadtgjacobianel{ij}\Big) R_i(\posjoint).
        \end{equation}
        The variable $\hat{B}_i^w$ from Equation \eqref{eq:ga_inertia_matrix_dt} can be found as 
        \begin{equation}\label{eq:ga_inertia_bivector_rotor}
            \hat{B}_i^w = \left( I_3B_i^w \right)  \outer \left( R_i(\posjoint) \mathcal{I}\left( \reverse{R}_i(\posjoint) I_3B_i^w R_i(\posjoint)  \right)  \reverse{R}_i(\posjoint) \right).
        \end{equation}
        Note that in this case $I_3$ stands for $\gae{123}$, which is the pseudoscalar of the Euclidean geometric algebra $\mathbb{G}_{3}$, which is a sub-algebra of CGA. $B_i^w$ on the other hand is the bivector velocity that results from multiplying the geometric Jacobian with the joint velocity, i.e. $B_i^w = \gamatrix{J}_i^G(\bm{q})\bm{\dot{q}}$. The quantity $I_3B_i^w$ therefore is the angular velocity and is non-zero in $\gae{1}$,$\gae{2}$ and $\gae{3}$. The outer product in Equation \eqref{eq:ga_inertia_bivector_rotor} causes the quantity $\hat{B}_i^w$ to be a bivector again, i.e. the elements $\gae{23}$,$\gae{13}$ and $\gae{12}$ are non-zero.

        In all the above equations $\mathcal{I}(\cdot)$ expresses the inertia tensor being applied to a multivector or to each multivector element in the matrix case. The inertia tensor is a grade-preserving operation since it maps bivectors to bivectors \cite{doranGeometricAlgebraPhysicists2003}.
        
        Finally, we find the manipulator inverse dynamics equation in geometric algebra to be
        \begin{multline}\label{eq:ga_manipulator_inverse_dynamics}
            \bm{\tau}(\posjoint,\veljoint,\accjoint) = \bm{\tau}_{ext} + \gamatrix{I}(\posjoint)\accjoint + \gamatrix{\dot{I}}(\posjoint,\veljoint)
            \\
            + \gamatrix{V}^\trsp(\posjoint)\bm{m}
            \Big(\gamatrix{V}(\posjoint) \accjoint + \gamatrix{\dot{V}}(\posjoint,\veljoint)\veljoint + \gamatrix{G}\Big),
        \end{multline}
        and consequently the forward dynamics of a serial manipulator can be expressed as 
        \begin{multline}\label{eq:ga_manipulator_forward_dynamics}
            \accjoint(\posjoint,\veljoint,\bm{\tau}) = 
            \Big(\gamatrix{I}(\posjoint) + \gamatrix{V}^\trsp(\posjoint)\bm{m}\gamatrix{V}(\posjoint) \Big)^{-1}
            \\
            \Big(\bm{\tau} - \bm{\tau}_{ext}
            - \gamatrix{V}^\trsp(\posjoint)\bm{m}
            \Big(\gamatrix{\dot{V}}(\posjoint,\veljoint)\veljoint + \gamatrix{G}\Big)
            - \gamatrix{\dot{I}}(\posjoint,\veljoint)\Big).
        \end{multline}

        In the experiments, we will validate these equations by employing an inverse dynamics control scheme on top of the MPC in order to convert the acceleration commands to torques.
    % subsection dynamics_of_serial_manipulators (end)
% section sec:algorithm (end)

% % % % % % % % % % % % % % % % % % % % % % % % % % % % % % % % % % % % % % % % % % % % 

\section{Geometric Algebra for Optimal Control}
\label{sec:geometric_algebra_for_optimal_control}
    In this section, we describe how geometric algebra can be used in optimal control problems. We first present a brief review of optimal control, then its application to a pointmass system and finally show how it can be used for serial manipulators. 

    % % % % % % % % % % % % % % % % % % % % % % % % % % % % % % % % % % % % % % % % % % % % 

    \subsection{Optimal Control}
    \label{sub:optimal_control}
        Optimal control is a well-known technique that deals with the problem of finding a control sequence that minimizes an objective function. This objective function encodes the requirements of the task as well as the constraints of the robot and the environment. Modelling these mathematically requires special care, since they will determine the quality of the resulting solution. Furthermore optimal control can be applied as solver to a model predictive control problem, which requires fast convergence in order to achieve acceptable real time control rates. We will show that the use of geometric algebra for geometric primitives improves the clarity of equations and thus reduces computational difficulties. The modelling of the cost functions becomes easier and is done uniformly across all different primitives and is done directly in the error vector as opposed to the precision matrix, which results in a low symbolic complexity of expressions and a geometric intuitiveness, i.e. geometric meaning can directly be inferred. 

        Discrete-time optimal control aims at finding a control sequence that minimizes the cost function 
        \begin{equation}\label{eq:oc_cost_function}
            \min_{\bm{u}} L (\bm{x},\bm{u}) 
            = l_f(\bm{x}_N) 
            + \sum_{k=1}^{N-1}
             l_k(\bm{x}_k)
              +\|\bm{u}\|_{R}^2,
        \end{equation}
        \textcolor{black}{where $l_k(\bm{x}_k)$ and $l_f(k\bm{x}_N)$ are the state dependent running and final cost, respectively, and $\|\bm{u}\|_R^2$ is a regularization term representing a control cost.}
        A popular method to solve this problem is the iterative Linear Quadratic Regulator (iLQR) \cite{tassaSynthesisStabilizationComplex2012}. It solves the problem by linearizing the non-linear system around the current solution and by assuming a quadratic cost. The solution is then refined iteratively until convergence. We will be using iLQR to solve the problem in a model predictive control (MPC) fashion in the experiments. This means that we are solving the regulation problem at each timestep and apply only the first control command to the robot. 
    % subtion sec:optimal_control (end)

    % % % % % % % % % % % % % % % % % % % % % % % % % % % % % % % % % % % % % % % %
    
    \subsection{Optimal Control on the Motor Manifold}
    \label{sub:optimal_control_on_the_motor_manifold}
        Employing homogeneous coordinates in 4D geometric algebra effectively allows us to linearize rigid body motions in 3D Euclidean space. In order to exploit this useful property, we demonstrate how to solve reaching tasks for rigid body motion using the motor manifold. 
        The linear system is defined in the linear 6-dimensional bivector space, i.e. the Lie algebra of the motor manifold. The state $\bm{x}$ is defined to be the stacked vector of the parameter vectors $\bm{b}$ and $\dot{\bm{b}}$ of the bivector $B$ and its time derivative $\dot{B}$, respectively. From this follows the definition of the linear dynamical system as 
        \begin{equation}\label{eq:bivector_linear_system}
            \bm{x}_{t+1} = \begin{bmatrix}\bm{b}_{t+1}\\ \dot{\bm{b}}_{t+1}\end{bmatrix} = \bm{A} \begin{bmatrix}\bm{b}_{t}\\ \dot{\bm{b}}_{t}\end{bmatrix} + \bm{C} \bm{u}_t,
        \end{equation}
        where the command $\bm{u}$ corresponds to bivector accelerations. 

        Using the optimal control formulation that was presented in Equation \eqref{eq:oc_cost_function}, we now need to define appropriate state costs based on geometric algebra. Naturally the inverse kinematics cost function that was presented in Section \ref{sub:inverse_kinematics} can be used in order to define pose targets for reaching motions. This cost function is therefore the most equivalent to classical methods using e.g. transformation matrices. Of course, when using this linear bivector system, the current motor $M_t$ is not found using the forward kinematics, but instead using the exponential map, i.e.
        \begin{equation}\label{eq:oc_current_motor}
            M_t = \exp(B(\bm{b}_t)).
        \end{equation}
        The corresponding cost function therefore becomes
        \begin{equation}\label{eq:bivector_motor_cost}
            l(\bm{x}_t) = \big\|\bm{e}(\bm{x}_t)\big\|_2^2 = \Big\| \log \left( \reverse{M}_{\text{target}} M_t \right) \Big\|_2^2.
        \end{equation}
        Note that due to the logarithmic map that is used here, the error vector $\bm{e}(\bm{x}_t)$ is the parameter vector of a bivector and is hence 6-dimensional as well. Since the motors include orientation as well as position, it essentially is an oriented pointmass system. 

        Apart from defining target poses using the motor manifold, geometric algebra additionally offers the possibility to define targets using its geometric primitives and the accompanying incidence relationships. We exploit the nullspace representations of the primitives for the formulation of the reaching objectives, more specifically we use the OPNS representation. By definition of the OPNS, the outer product is zero for any point that is on a geometric primitive. The multivector valued error can therefore be defined as
        \begin{equation}\label{eq:ga_oc_cost_function}
            E(\posjoint) = X_d \outer M_t X \reverse{M}_t,
        \end{equation}
        with $X = \gae{0}$, i.e. the point at the origin, in this case and $X_d$ can be any geometric primitive that can be expressed in the algebra. It is important to highlight that other combinations are possible as well and $X$ is not restricted to be a point, it can for example also be a line with $X_d$ being a point, which will be shown later in the form of a pointing task for a manipulator. Note that the structure of equation remains the same regardless of the combination of primitives.
    % % % % % % % % % % % % % % % % % % % % % % % % % % % % % % % % % % % % % % % % % %
    
    \subsection{Optimal Control for Serial Manipulators}
    \label{sub:optimal_control_for_serial_manipulators}
        In this section, we are presenting the formulation of objective functions for optimal control problems with serial manipulators based on geometric algebra. Similarly to the previous section, the inverse kinematics cost function can be used to define target poses for a manipulator to reach. More interesting is to consider Equation \eqref{eq:ga_oc_cost_function} for manipulators, of course in this case the motor again corresponds to the forward kinematics function. Using $X = \gae{0}$ therefore means that the expression  $M(\posjoint) X \reverse{M}(\posjoint)$ corresponds to the tip of the end-effector. $X_d$ again is free to be any geometric primitive. In all cases the Jacobian can be found by applying the chain rule to the multivector expressions
        \begin{equation}\label{eq:ga_cost_function_jacobian}
            \gamatrix{J}^E(\posjoint) = X_d \outer \left( \gaajacobian X \reverse{M}(\posjoint) + M(\posjoint) X \reverse{\gamatrix{J}^A}(\posjoint)\right),
        \end{equation}
        where the $\gaajacobian$ is the analytic Jacobian that was presented in Equation \eqref{eq:ga_analytic_jacobian} and the reverse of a multivector matrix is defined as the element-wise multivector reverse.
        
        Of course, depending on the combination of $X$ and $X_d$, the resulting $E(\posjoint)$ will represent a different geometric meaning, which can be seen by the different non-trivial blades it holds. It is, however, known a priori what the resulting non-trivial blades are. From this it follows that the embedding function $\mathcal{E}$ actually becomes dependent on $X$ and $X_d$, i.e.
        \begin{equation}\label{eq:cost_function_jacobian_embedding}
            \bm{J}^E(\posjoint) = \mathcal{E}(X,X_d)\left[ \gamatrix{J}^E(\posjoint) \right].
        \end{equation}
        The purpose of the embedding function thus is the removal of the trivial blades of the multivector, i.e. removing the zero rows from the matrix. This is in line with the goal of keeping the representations compact to allow for efficient computation. Furthermore, the embedding now allows the usage of off-the-shelf tools for optimal control.

        Note that in general no special cases, such as division by zero, need to be considered here. The exception-free incidence property of conformal geometric algebra allows for the equations to be coded exactly as they are presented in this paper. 
    % subsection optimal_control_for_serial_manipulators (end)
% section sec:geometric_algebra_for_optimal_control (end)

% % % % % % % % % % % % % % % % % % % % % % % % % % % % % % % % % % % % % % % % % % % % 

\section{Experiments}
\label{sec:experiments}
    In this section, we are presenting implementation details of the provided library \textit{gafro} as well as benchmarks of the kinematics computation. Afterwards we show various experiments with the Franka Emika robot to demonstrate how geometric algebra can used to model different tasks.

    % % % % % % % % % % % % % % % % % % % % % % % % % % % % % % % % % % % % % % % % % %
    
    \subsection{Implementation Details}
    \label{sub:implementation_details}
        We implemented the presented robotics kinematics and dynamics algorithms along with cost functions for optimal in control in c++20. This resulted in the library \textit{gafro}, which is publicly available. In this section, we are presenting this library on a high-level and highlight some of its features. A more in-depth presentation, exhaustive benchmarks and comparison to other libraries will be part of future work.

        At the core of \textit{gafro} is a custom implementation of conformal geometric algebra. It exploits the sparsity of the multivector by only storing the data blades that are non-zero by the structure of the objects. The geometric, inner and outer products are implemented as expression templates, that are further exploiting this structure by only evaluating the elements of the resulting type that are known to be non-zero. The types are evaluated at compile time and the evaluation tree is constructed, which is then evaluated at runtime in a lazy fashion. One of our design goals for the library was the seamless integration with existing tools for robotics such as libraries for optimization and optimal control. To this end, we used the Eigen library \footnote{\url{https://eigen.tuxfamily.org}}, which is de facto the standard tool in robotics, to implement the sparse parameter vector of the multivectors.  

\begin{figure}[!ht]
            % % % %
            \centering
            % % % % 
            % \begin{subfigure}[t]{\linewidth}
            %     \includegraphics[width=\linewidth]{images/benchmarks/benchmark_forward_kinematics.pdf}
            %     \caption{Results for benchmarking the forward kinematics. The reference robotic system is the Franka Emika robot.}
            % \end{subfigure}
            % % % % % 
            % \begin{subfigure}[t]{\linewidth}
            %     \includegraphics[width=\linewidth]{images/benchmarks/benchmark_geometric_jacobian.pdf}
            %     \caption{Results for benchmarking the geometric Jacobian.}
            % \end{subfigure}
            \includegraphics[width=\linewidth]{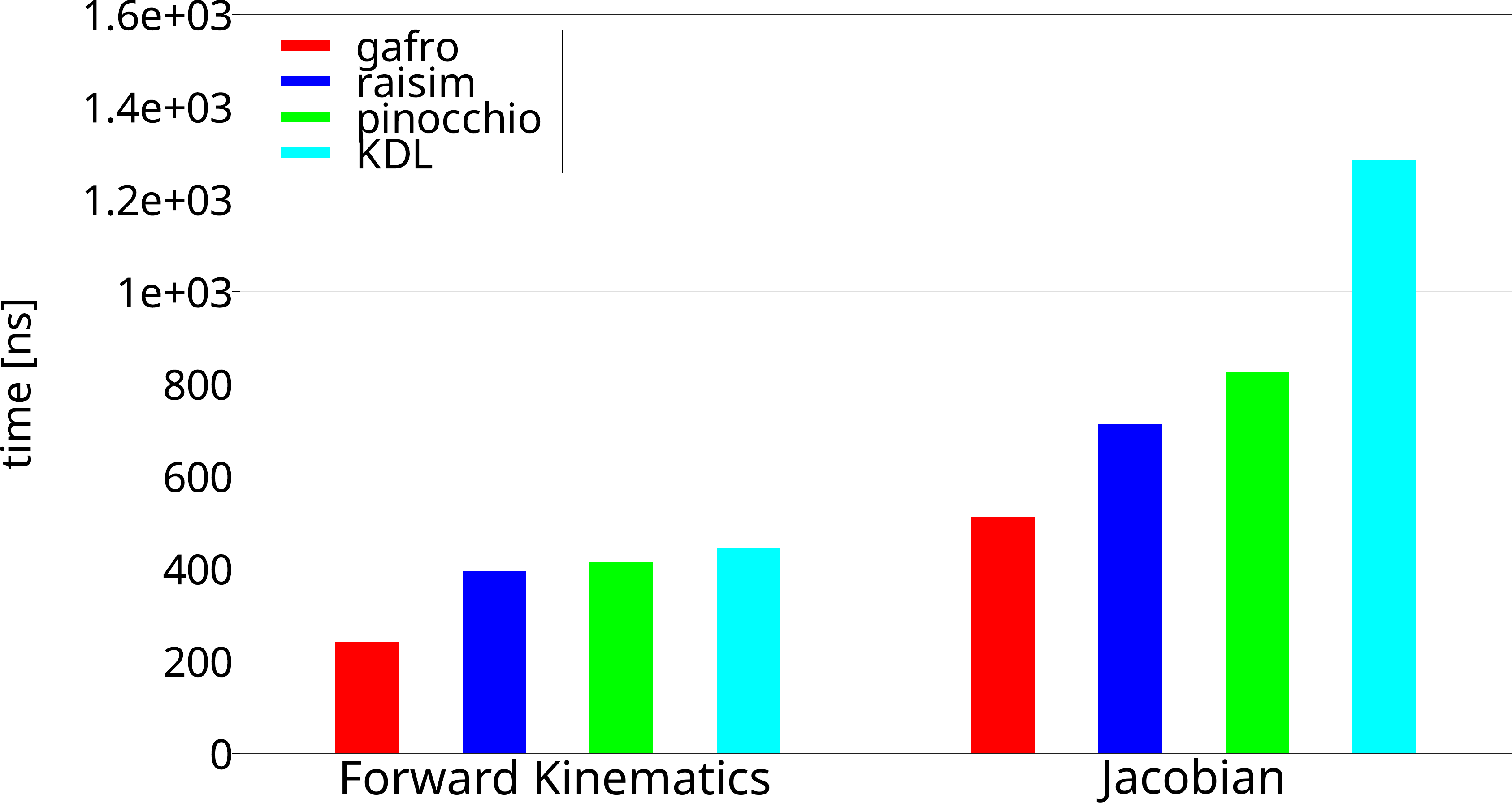}
            % % % % 
            \caption{Benchmarking results for \textit{gafro} compared to Raisim, Pinocchio and KDL. The benchmarks were all performed on an AMD Ryzen 7 4800U CPU using the compiler flags \texttt{-O3 -msse3 -march=native}. The presented results are the average of 10000 executions with 10 repetitions.}
            % % % % 
            \label{fig:benchmarking_results_of_the_geometric_jacobian}
        \end{figure}

        Since this library implements robot kinematics and dynamics algorithms, we are comparing and benchmarking \textit{gafro} against several libraries that are commonly used in robotics applications. These libraries include Raisim \cite{hwangboPerContactIterationMethod2018}, Pinocchio \cite{carpentierPinocchioLibraryFast2019} and KDL \cite{smitsKDLKinematicsDynamics}. An excerpt of the benchmarking results can be found in Figure \ref{fig:benchmarking_results_of_the_geometric_jacobian}. As can be seen, our library can compute these important quantities considerably faster than the other libraries that are based on homogeneous transformation matrices. \textcolor{black}{Note that previous publications have already shown the advantages that dual quaternions have over homogeneous transformation matrices \cite{dantamRobustEfficientForward2021} in terms of computational complexity. Due to the close relationship that CGA motors have with dual quaternions, the same advantages apply to them as well. In \cite{dantamRobustEfficientForward2021} a 30\%-40\% improvement in performance of dual quaternions compared to homogeneous transformation matrices was reported, which is similar to our findings for motors.}

        % % % %

        We want to point out that these benchmarks are preliminary results only, since there are some code optimizations that still need to be done. This especially applies to the computation of the dynamics that was presented in this paper. While it is possible to do real-time control with our implementation, it is still naive in the sense that algorithmic improvements of the implementation will make the computation faster. Both these issues will be addressed in future work that will be dedicated to the implementation details as well as an in-depth benchmarking against a wider range of libraries. At this point we will then also provide python bindings for the library. 
    % subsection implementation_details (end)

    % % % % % % % % % % % % % % % % % % % % % % % % % % % % % % % % % % % % % % % % % % % % 

    \subsection{Torque Control of Serial Manipulators}
    \label{sub:torque_control_of_serial_manipulators}
        The optimal control methods derived in Section \ref{sub:optimal_control} are used in a model predictive control framework. In order to achieve fast, online computation we are employing a double integrator system in the joint space, which results in acceleration commands for the control of the manipulator. However standard practice is using torque commands, which means that we have to convert the accelerations to torques. This can be realized using an inverse dynamics controller, the required control command can thus be found as 
        \begin{equation}\label{eq:inverse_dynamics_control}
            \bm{u}_{\tau} =  \bm{\tau}(\bm{q}, \bm{\dot{q}}, \bm{\ddot{q}}_d) + \bm{K}_p (\bm{q}_d - \bm{q})+ \bm{K}_d (\bm{\dot{q}}_d - \bm{\dot{q}}),
        \end{equation}
        where the torque vector $\bm{\tau}$ is computed as presented in Equation \eqref{eq:ga_manipulator_inverse_dynamics}. $\bm{K}_p$ and $\bm{K}_d$ are the stiffness and damping gains, respectively.
        
        While the formulation of the inverse dynamics controller is following standard practice, the computation of the torques $\bm{\tau}(\bm{q}, \bm{\dot{q}}, \bm{\ddot{q}}_d)$ is done using the geometric algebra approach that was presented in Section \ref{sub:dynamics_of_serial_manipulators}. The experiments on the real robot therefore not only show GA can be used for tracking different geometric primitives online with MPC, it also verifies the dynamics computation in GA. Most importantly it validates the non-trivial inertia matrix that we derived in this paper.
    % section sec:torque_control_of_serial_manipulators (end)

    % % % % % % % % % % % % % % % % % % % % % % % % % % % % % % % % % % % % % % % % % %
    
    \subsection{Inverse Kinematics}
    \label{sub:inverse_kinematics}
        In order to evaluate the numerical inverse kinematics using the motor formulation, we repeated the following experiment 10000 times. We sampled a random target from within the workspace of the Franka Emika robot and an initial joint configuration. Then using the standard Gauss-Newton approach that we also described in Section \ref{sub:inverse_kinematics_of_serial_manipulators}, we computed the optimal solution. \textcolor{black}{The solver success rate was 85.39\% with a tolerance of $1e^{-6}$.} The resulting final cost was in the order of $1\times10^{-10}$ on average and was found within 11.2 iterations, which corresponds to a time of 79 $\mu s$ on our system. \textcolor{black}{We considered only the successul solves for these statistics.} \textcolor{black}{Note that this inverse kinematics solution presented in this work is meant as an explanatory example and proof of concept. It is not meant to compete in this form with existing IK solvers, but rather it should motivate to integrate GA into them. A potential future work could therefore be augmenting state-of-the-art solvers like TRAC-IK \cite{beesonTRACIKOpensourceLibrary2015} with GA. Currently TRAC-IK uses KDL in its implementation to calculate the forward kinematic chain. Based on our benchmarks, using \textit{gafro} instead of KDL would lead to a significant increase in performance.}
    % subsection inverse_kinematics (end)

    % % % % % % % % % % % % % % % % % % % % % % % % % % % % % % % % % % % % % % % % % %
    
    \subsection{Pointmass System}
    \label{sub:pointmass_system}
\begin{figure}[!ht]
            % % % %
            \centering
            % % % % 
            \includegraphics[width=0.85\linewidth]{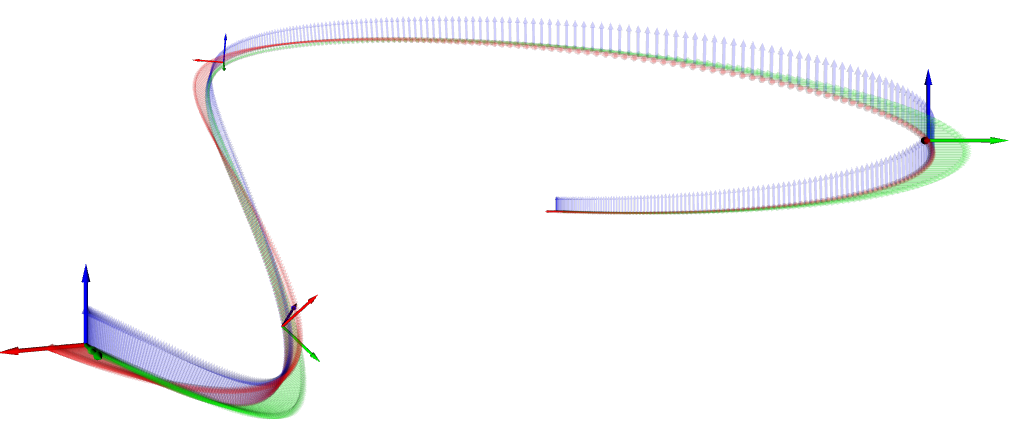}
            % % % % 
            \caption{Optimal trajectory for an oriented pointmass system. We placed four target motors along the trajectory at $T/4$, $T/2$, $3T/4$ and $T$, respectively. The target motors are highlighted along the trajectory. The optimal trajectory was then found using the system defined in Equation \eqref{eq:bivector_linear_system} and the cost function from Equation \eqref{eq:motor_ik_manifold}.}
            % % % % 
            \label{fig:oriented_pointmass_system_}
        \end{figure}

        In Sections \ref{sub:inverse_kinematics_of_serial_manipulators} and \ref{sub:optimal_control_on_the_motor_manifold} we first presented the cost function to minimize the difference between two motors and then optimal control for an oriented pointmass formulated as a linear system in the bivector space. Here we present an optimal trajectory for such a system using several target motors. This example of a control problem using several target motors along the trajectory is shown in Figure \ref{fig:oriented_pointmass_system_}.
        % % % %

        Note that the same objective can be formulated for manipulators as well, which would correspond to reaching target poses with the end-effector, which makes it similar to classical methods, albeit with a different mathematical formulation. Therefore we omit showing it for manipulators for brevity and concentrate on modeling and reaching tasks using the geometric primitives in the following sections.
    % subsection pointmass_system (end)

    % % % % % % % % % % % % % % % % % % % % % % % % % % % % % % % % % % % % % % % % % %
    
    \subsection{Reaching Tasks}
    \label{sub:target_point_reaching}
\begin{figure}[!ht]
            % % % %
            \centering
            % % % % 
            \begin{subfigure}[]{0.49\linewidth}
                \centering
                \includegraphics[width=0.75\linewidth]{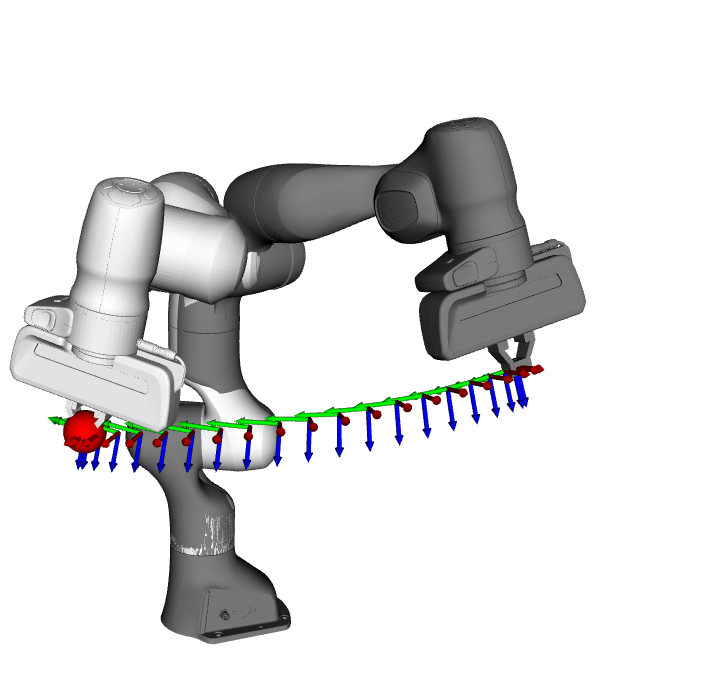}
                \caption{Reaching a point.}
                \label{fig:experiment_reaching_a_point}
            \end{subfigure}
            % % % % 
            \begin{subfigure}[]{0.49\linewidth}
                \centering
                \includegraphics[width=0.75\linewidth]{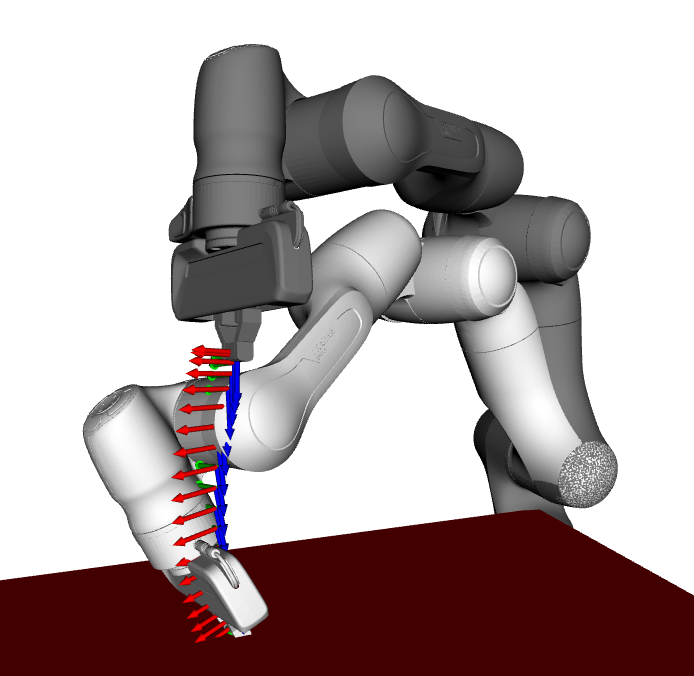}
                \caption{Reaching a plane.}
                \label{fig:experiment_reaching_a_plane}
            \end{subfigure}
            % % % % 
            \begin{subfigure}[]{0.49\linewidth}
                \centering
                \includegraphics[width=0.75\linewidth]{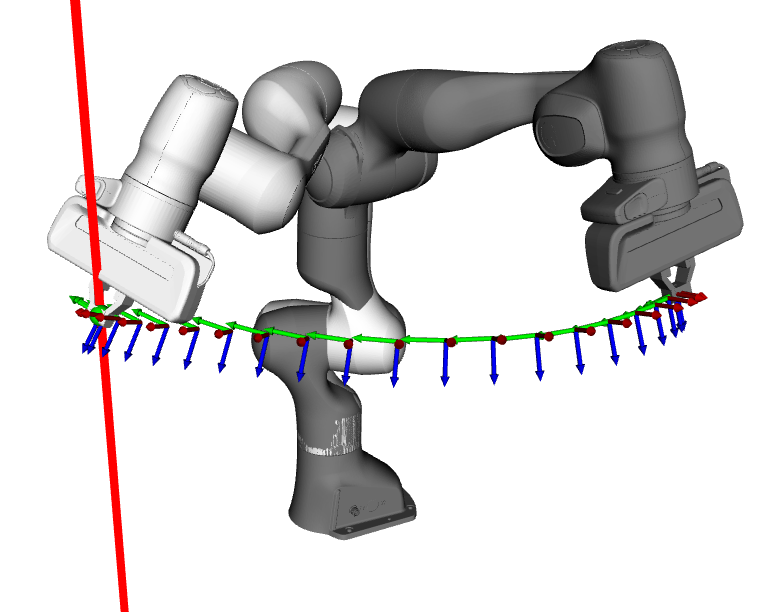}
                \caption{Reaching a line.}
                \label{fig:experiment_reaching_a_line}
            \end{subfigure}
            % % % % 
            \begin{subfigure}[]{0.49\linewidth}
                \centering
                \includegraphics[width=0.75\linewidth]{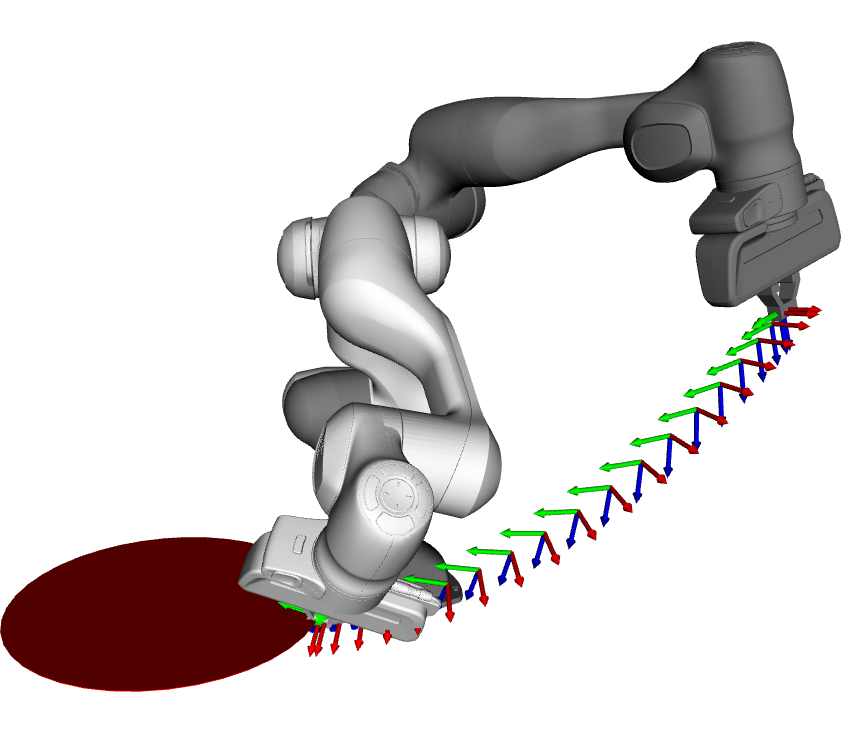}
                \caption{Reaching a circle.}
                \label{fig:experiment_reaching_a_circle}
            \end{subfigure}
            % % % % 
            \caption{Examples of optimal trajectories for reaching tasks using different geometric primitives. The initial configuration is always shown in gray and the final one in white. The target geometric primitive is shown in red. And the trajectory is depicted as the frames corresponding to the end-effector.}
            % % % % 
            \label{fig:experimental_setup_for_the_reaching_tasks_}
        \end{figure}

        Using the cost function formulation of geometric algebra that was presented in Equation \eqref{eq:ga_oc_cost_function} various reaching tasks can be defined. In general, for a reaching task, the end-effector should reach a certain position. This can be modeled by using a point for $X$. Then the desired multivector $X_d$ can be any other geometric primitive, which in turn means that instead of only reaching a point, we can also reach lines, planes, circles and spheres. Higher order quadrics are possible as well, this however remains the subject of further investigations. We present optimal trajectories that were computed using the iterative linear quadratic regulator to explain how different geometric primitives can be reached using the same structure of the cost function, which is shown in Figure \ref{fig:experimental_setup_for_the_reaching_tasks_}. In the experiments using the real Franka Emika we are then using nominal MPC, which results in an offset for the steady-state. This effect is expected but negligible in our work, since the focus of this paper are the modeling aspects. Since we are using an MPC framework we do not need to use fixed points for the reaching but can have movable targets. In practice we use Aruco markers to track the target online or we are disturbing the robot while it is moving. 

        % % % %

        % % % % % % % % % % % % % % % % % % % % % % % % % % % % % % % % % % % % % % % %
        
        \subsubsection{Reaching a Point}
        \label{ssub:reaching_a_point}
            Reaching a point means that the desired geometric primitive is a point, i.e. $X_d = P$ in Equation \eqref{eq:ga_oc_cost_function}. The reference primitive $X_r$ is a point as well and represents the tip of the end-effector. Figure \ref{fig:experiment_reaching_a_point} shows an example trajectory for reaching a fixed point from a random initial configuration.

\begin{figure}[!ht]
                % % % %
                \centering
                % % % % 
                \begin{subfigure}[]{\linewidth}
                    \centering
                    \includegraphics[width=\linewidth]{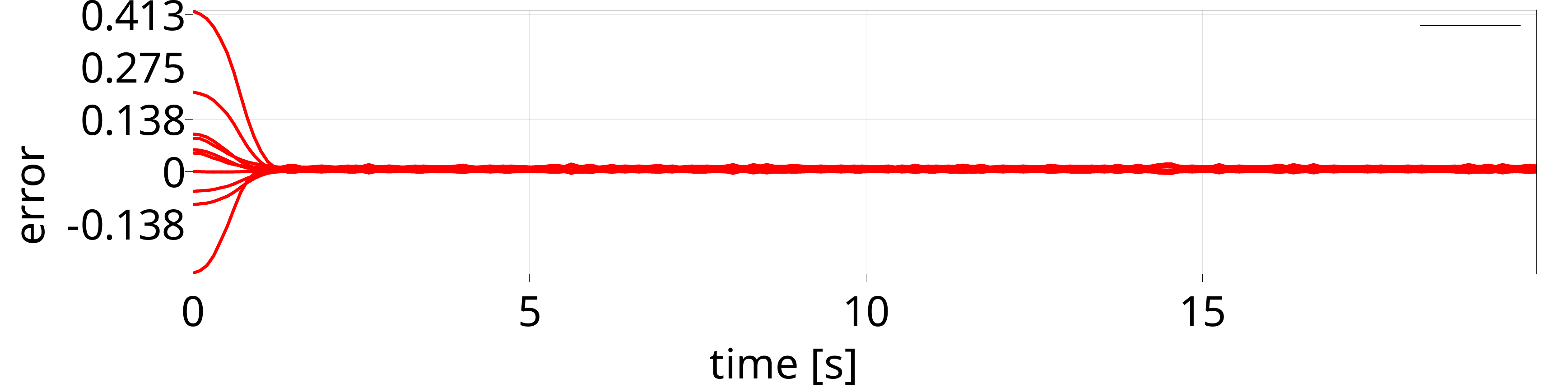}
                    \caption{Regulation of the end-effector to a target point using MPC without disturbing it. It can be seen that the offset that is induced by the nominal MPC is only very small.}
                    \label{fig:experiment_mpc_point_error}
                \end{subfigure}
                \begin{subfigure}[]{\linewidth}
                    \centering
                    \includegraphics[width=\linewidth]{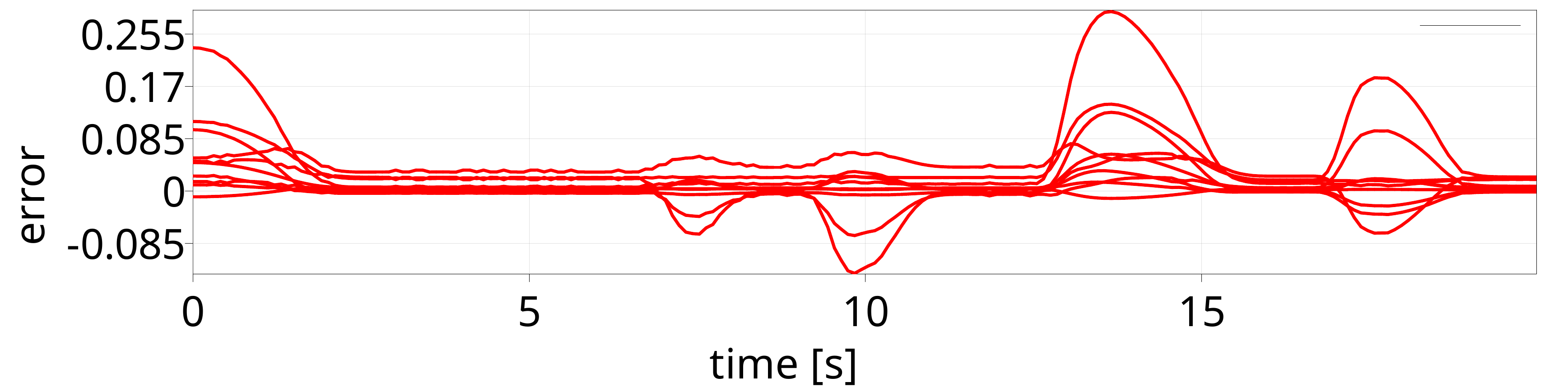}
                    \caption{Regulation of the end-effector to a target point using MPC while disturbing it.}
                    \label{fig:experiment_mpc_point_error_disturbance}
                \end{subfigure}
                % % % % 
                \caption{Error of reaching a fixed target point in MPC for a total duration of 20s. The individual lines show the elements of resulting 10-dimensional error vector that results from the outer product of two points.}
                % % % % 
                \label{fig:error_of_reaching_a_target_point_in_mpc_}
            \end{figure}

            In the real robot experiments we used a single Aruco marker for reaching a movable point. For safety reasons the target point was set 10cm above the marker. We then moved the marker around allowing the robot to follow the reference. We present the results of the real experiment in Figure \ref{fig:error_of_reaching_a_target_point_in_mpc_}. In both plots that are presented we show the values of the 10-dimensional error vector that results from the outer product of two points. If the magnitude of this vector is zero it means that the reference point is in the nullspace of the desired point w.r.t to the outer product. In the case of points, this means that they are identical and the target is reached. Figure \ref{fig:experiment_mpc_point_error} presents the static case where we neither moved the point nor disturbed the robot while it moved. We did both of these in the plot shown in Figure \ref{fig:experiment_mpc_point_error_disturbance}. It can be seen that the MPC controller is fast and reactive, reaching the targets in a stable manner.

            % % % %

            Reaching a point is in this context a trivial example, since it can be easily done using classical methods as well, but it serves to show that reaching problems can be solved for all geometric primitives in the same way as they are solved for a point using geometric algebra. 
        % subsubsection ssub:reaching_a_point (end)

        % % % % % % % % % % % % % % % % % % % % % % % % % % % % % % % % % % % % % % % %
        
        \subsubsection{Reaching a Pointpair}
        \label{ssub:reaching_a_pointpair}
            Using a pointpair as the target presents a special opportunity to model a control problem with options. A pointpair is the result of the outer product of two points. From this outer product nullspace representation, it follows that the outer product of the point pair and any point $P=\mathcal{C}(\bm{x})$ with $\bm{x}\in \mathbb{R}^3$ is zero if and only if $P$ is identical to one of the points that constructed the pointpair.

\begin{figure}[!ht]
                % % % %
                \centering
                % % % % 
                % % % % 
                \begin{subfigure}[]{0.49\linewidth}
                    \centering
                    \includegraphics[width=\linewidth]{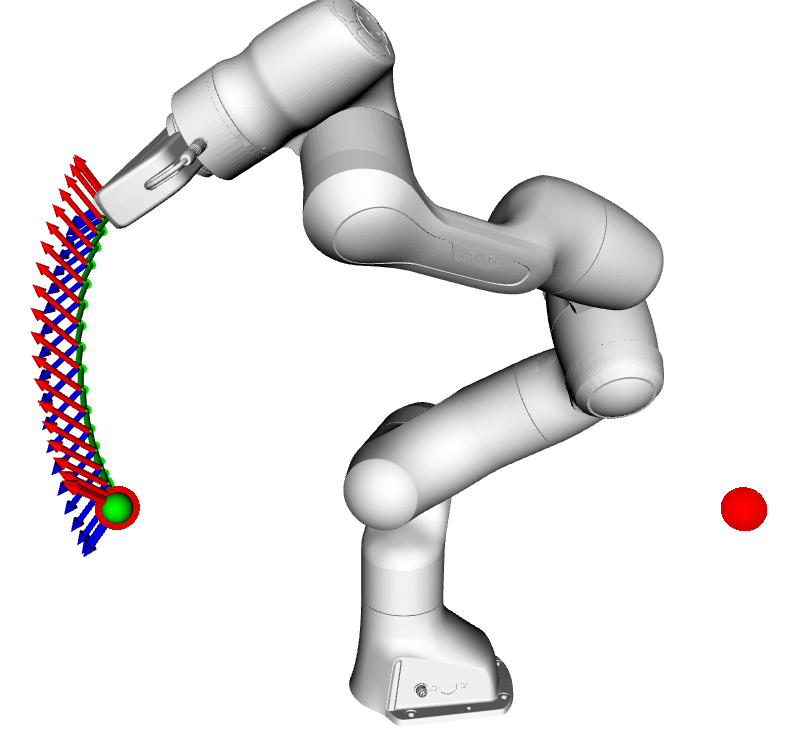}
                    \caption{Left}
                    \label{fig:reaching_a_pointpair_left}
                \end{subfigure}
                \hfill 
                \begin{subfigure}[]{0.49\linewidth}
                    \centering
                    \includegraphics[width=\linewidth]{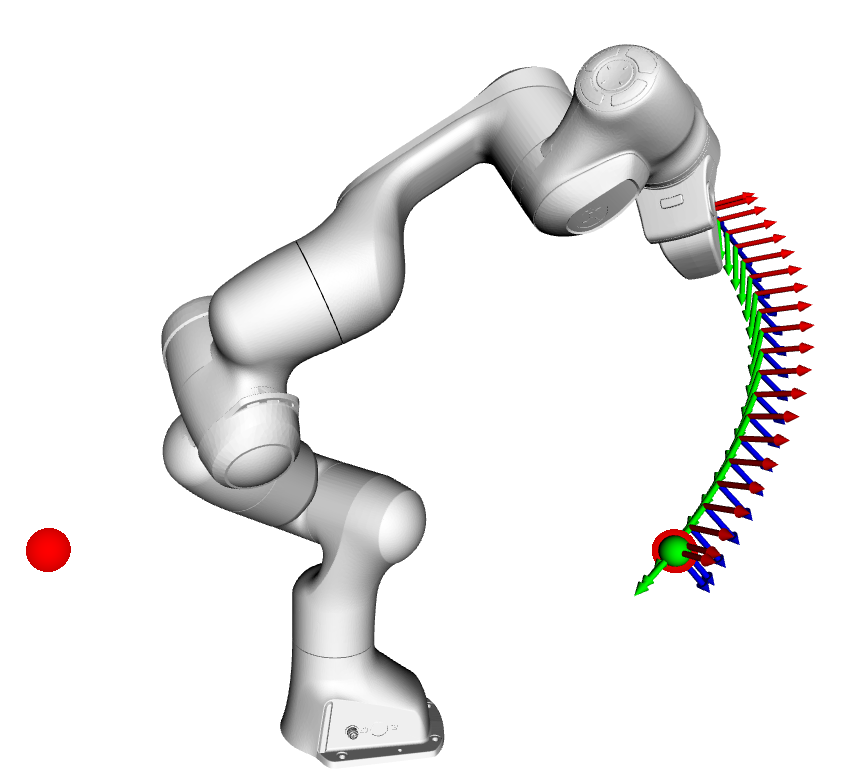}
                    \caption{Right}
                    \label{fig:reaching_a_pointpair_right}
                \end{subfigure}
                % % % % 
                \caption{Depending on the initial configuration, either the left or the right point of the pointpair is reached.}
                % % % % 
                \label{fig:reaching_a_pointpair}
            \end{figure}

            The two possibilities are shown in Figure \ref{fig:reaching_a_pointpair}, where Figure \ref{fig:reaching_a_pointpair_left} shows the robot reaching the first point and Figure \ref{fig:reaching_a_pointpair_right} the second one. The point that is reached depends on the initial configuration and there are no conditional statements required. The corresponding Jacobian is thus always computed in the same way, i.e. as presented in Equation \eqref{eq:ga_cost_function_jacobian}, and is valid without exceptions. The same is true for the other geometric primitives that we are considering here, but we wanted to specifically highlight the pointpair primitive due to its power to model a binary target. \textcolor{black}{We also want to point out here that the pointpair primitive would be very suitable for modeling dual-arm manipulation tasks. In this scenario the pointpair would represent the end-effector positions of the two manipulators.}
            % % % %
        % subsubsection ssub:reaching_a_pointpair (end)

        % % % % % % % % % % % % % % % % % % % % % % % % % % % % % % % % % % % % % % % %
        
        \subsubsection{Reaching a Plane}
        \label{ssub:reaching_a_plane}
            A plane in geometric algebra is represented by three individual points and $\gae{\infty}$ using the outer product nullspace, i.e. $X_d = E = P_1 \outer P_2 \outer P_3 \outer \gae{\infty}$. Note that we do not need to know the orientation, i.e. its normal vector, of the plane in order to define it. It is sufficient to know three points that lie in the plane. When multiplying the plane with a point using the outer product, any point that lies in the plane will result in zero, $E \outer P = 0$ if $P\in E$. Equation \eqref{eq:ga_oc_cost_function} will therefore minimize the reaching motion to the plane from any random initial configuration as shown in Figure \ref{fig:experiment_reaching_a_plane}.
        % subsubsection ssub:reaching_a_plane (end)

        % % % % % % % % % % % % % % % % % % % % % % % % % % % % % % % % % % % % % % % %
        
        \subsubsection{Reaching a Line}
        \label{ssub:reaching_a_line}
\begin{figure}[!ht]
                % % % %
                \centering
                % % % % 
                \includegraphics[width=\linewidth]{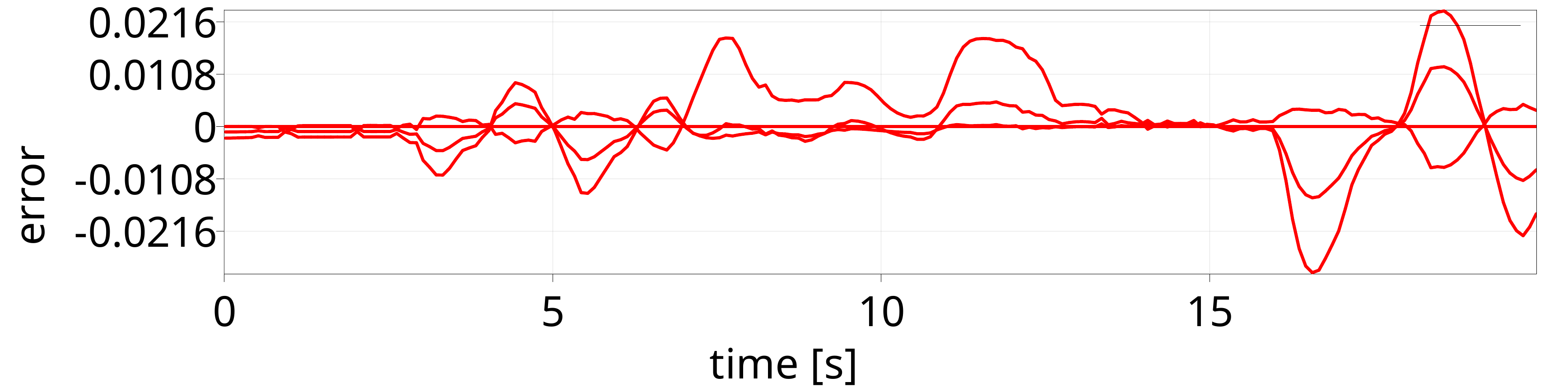}
                % % % % 
                \caption{Components of the error vector resulting from the outer product of a line and a point. Error of reaching a fixed target point in MPC for a total duration of 20s. The individual lines show the resulting error vector that results from the outer product of a line and a point.}
                % % % % 
                \label{fig:reaching_a_target_line_in_mpc}
            \end{figure}

            A line is similar to a plane, but requires only two known points along the line in order to construct it. We show an optimal trajectory for reaching a line in Figure \ref{fig:experiment_reaching_a_line}. For the real robot experiment we again used an Aruco marker, in this case one construction point was on the marker and the other one was 10cm above it. The target line therefore always is perpendicular to the $y-z-$plane. In Figure \ref{fig:reaching_a_target_line_in_mpc} we show the results of the experiment. The corresponding error vector has 6 components and is geometrically equivalent to a circle. 
            % % % %
        % subsubsection ssub:reaching_a_line (end)

        % % % % % % % % % % % % % % % % % % % % % % % % % % % % % % % % % % % % % % % %
        
        \subsubsection{Reaching a Circle}
        \label{ssub:reaching_a_circle}
            A target circle is constructed by the outer product of three points, i.e. $X_d = C = P_1 \outer P_2 \outer P_3$. These three points uniquely define the circle and no further knowledge about its radius or orientation is required (but both of these can of course be obtained from the circle for the visualization shown in Figure \ref{fig:experiment_reaching_a_circle}). The target is here only the boundary of circle (not the full disc).
        % subsubsection ssub:reaching_a_circle (end)
    % subsection target_point_reaching (end)
    
    \subsection{Pointing Task}
    \label{sub:pointing_task}
\begin{figure}[!ht]
            % % % %
            \centering
            % % % % 
            \includegraphics[width=0.5\linewidth]{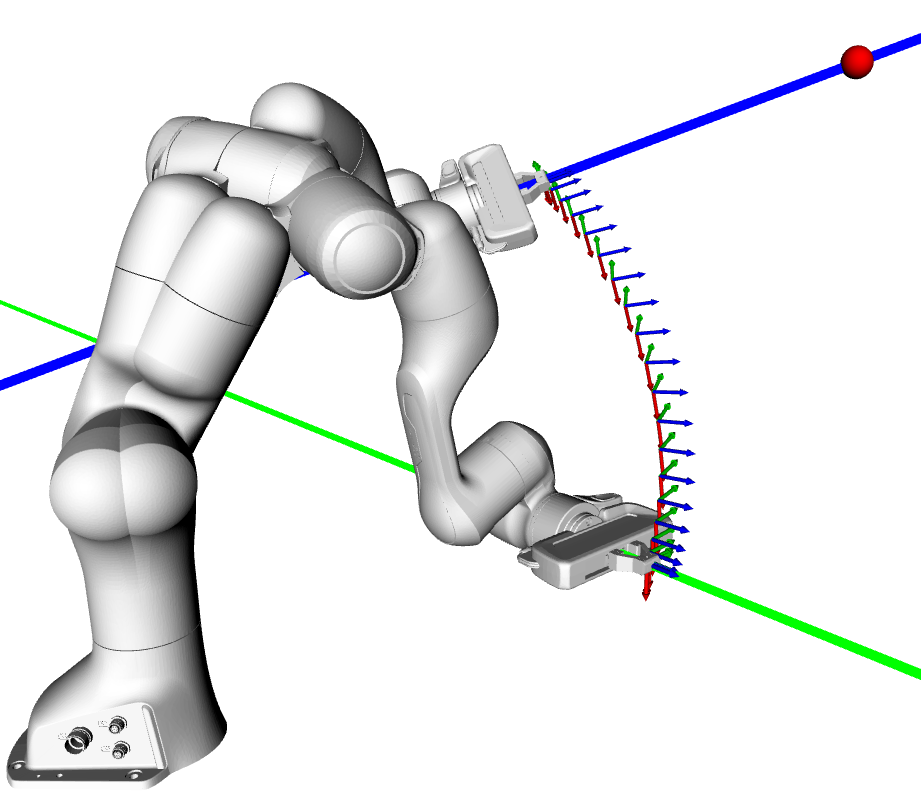}
            % % % % 
            \caption{optimal trajectory example for a pointing task. The target is shown as the red point. The pointing line is defined to be collinear to the $z$-axis of the end-effector frame. It is shown in green for the initial configuration and in blue for the final configuration.}
            % % % % 
            \label{fig:pointing_task_setup}
        \end{figure}

\begin{figure}[!ht]
            % % % %
            \centering
            % % % %
            \includegraphics[width=\linewidth]{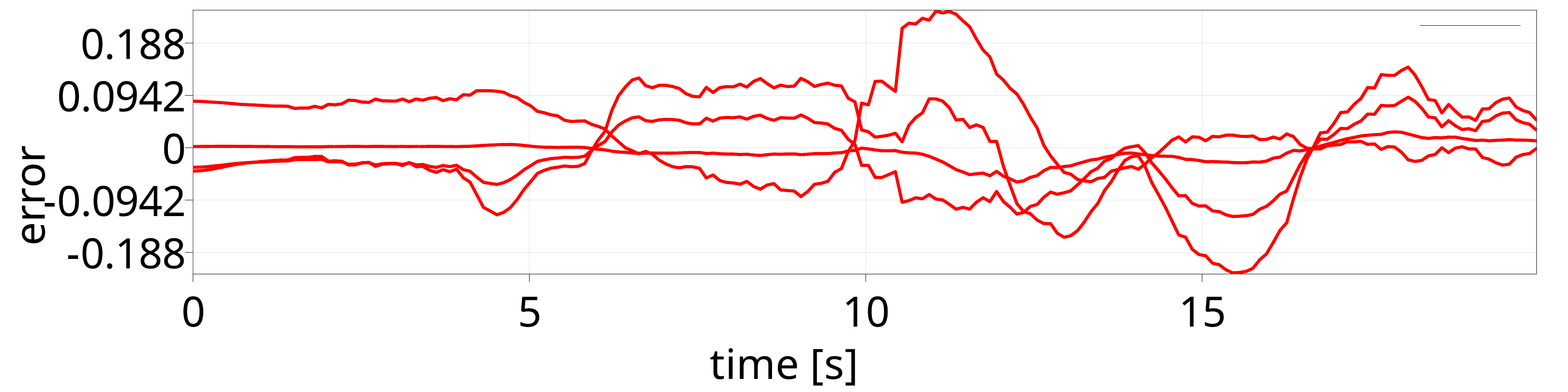}
            % % % % 
            \caption{Experimental results of the pointing task showing the components of the error vector during 20s. The Aruco marker representing the target point was constantly moved around, which is the reason why the error vector is more jittery than in other experiments.}
            % % % % 
            \label{fig:pointing_task_error}
        \end{figure}

        The modelling of a pointing task only requires the usage of a line instead of a point for $X$ in Equation \eqref{eq:ga_oc_cost_function}. A possible scenario where this task would be applied is tracking an object with a robot arm endowed with a camera. The line can in this case be interpreted as the line of sight of the camera. Again different geometric primitives can be used as the target, since the intersection of a line with any other primitive can be calculated in closed form without exceptions. The setup is shown in Figure \ref{fig:pointing_task_setup}. Figure \ref{fig:pointing_task_error} shows the moving target.

        % % % %
        % % % %
    % subsection pointing_task (end)

    % % % % % % % % % % % % % % % % % % % % % % % % % % % % % % % % % % % % % % % % % %
    
    \subsection{Circular Object Grasping Task}
    \label{sub:object_grasping_task}
\begin{figure}[!ht]
            % % % %
            \centering
            % % % % 
            \begin{subfigure}[]{\linewidth}
                \centering
                % % % %
                \begin{minipage}{0.49\linewidth} 
                    \centering
                    \raisebox{-\totalheight}{\includegraphics[width=\linewidth]{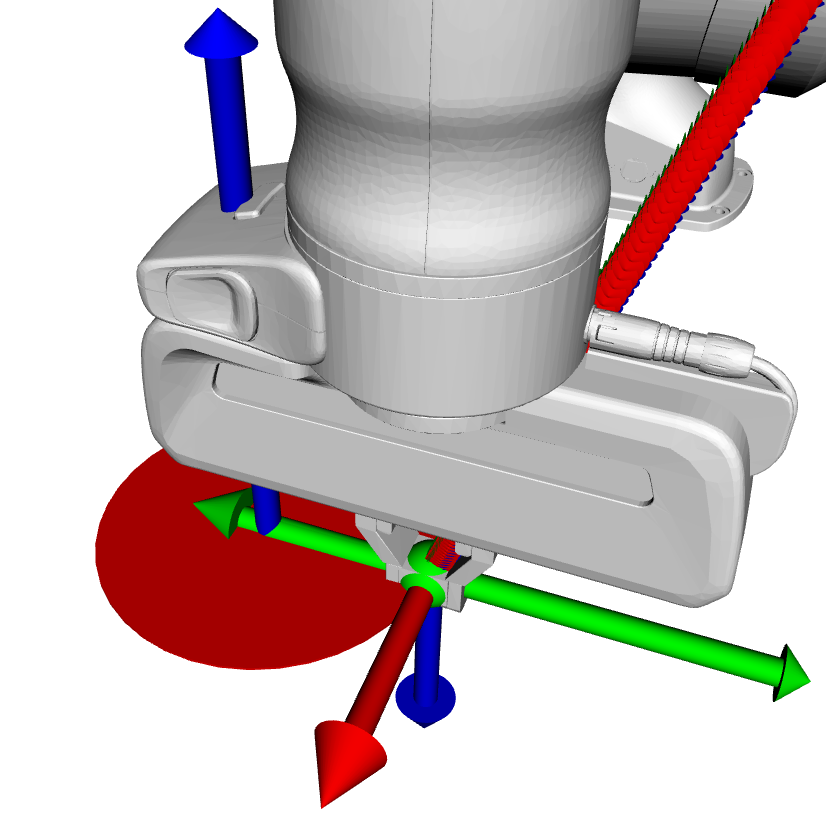}}
                \end{minipage}
                % % % % 
                \hfill
                % % % % 
                \begin{minipage}{0.49\linewidth}
                    \centering
                    \caption{Constraints defining the circular object grasping task: 1) the green point representing the end-effector position needs to lie on the circle (i.e. the boundary of the red disc), 2) the green arrows representing the $y$-axis of the end-effector frame and the radial vector of the circle must be collinear, 3) the blue arrows representing the $z$-axis of the end-effector frame and the normal vector of the circle must be collinear \underline{and} pointing in opposite directions.}
                \label{fig:grasping_task_constraints}
                \end{minipage}
            \end{subfigure}
            % % % % 
            \begin{subfigure}[]{\linewidth}
                \centering
                % % % %
                \begin{minipage}{0.49\linewidth} 
                    \centering
                    \caption{Franka Emika robot grasping a box with a circular opening. An Aruco marker is attached to the box to mark its location. The three points defining the circular opening are measured with respect to the marker frame.}
                \end{minipage}
                % % % % 
                \hfill
                % % % % 
                \begin{minipage}{0.49\linewidth}
                    \centering
                    \includegraphics[width=0.5\linewidth]{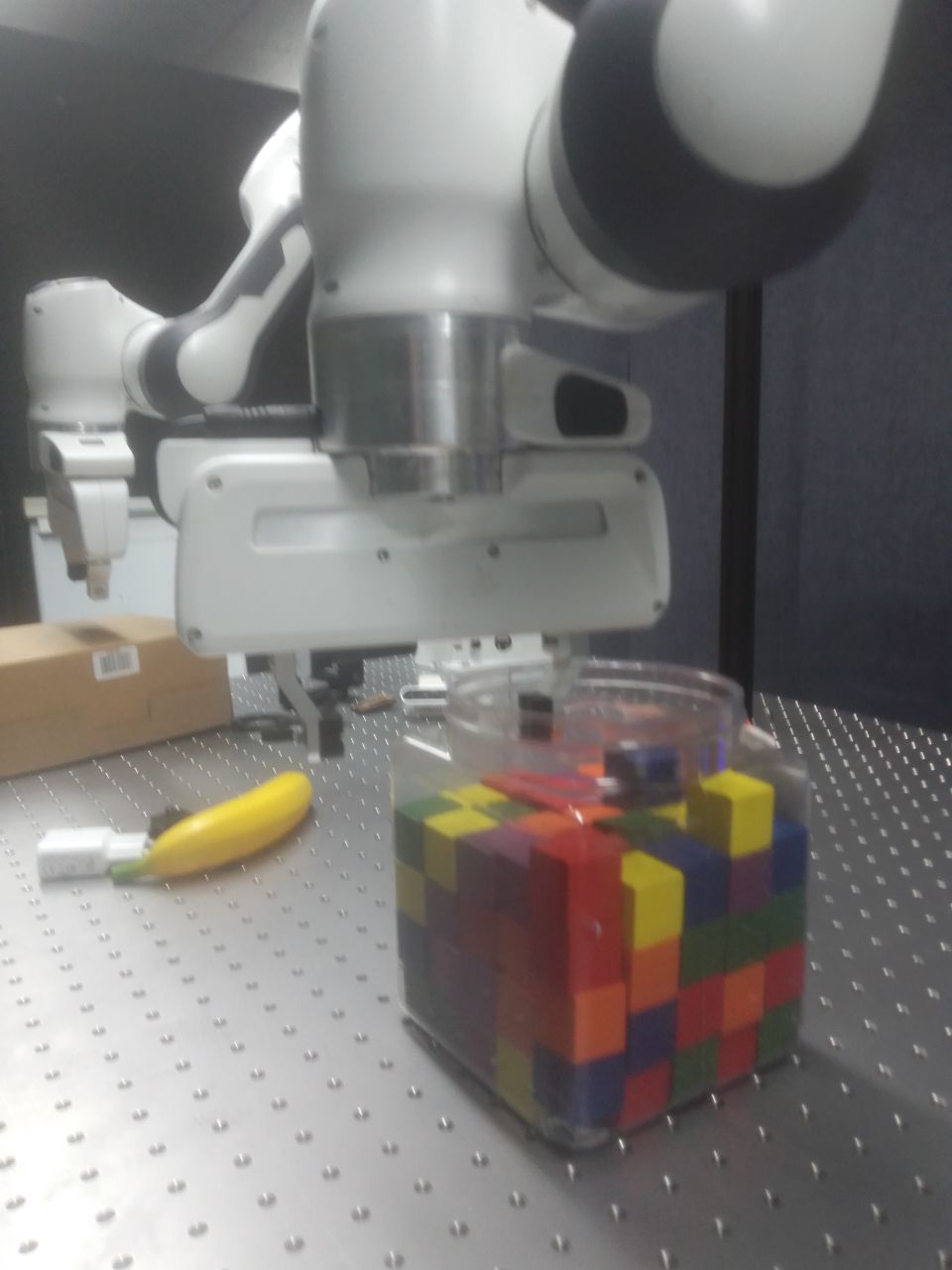}
                \end{minipage}
                \label{fig:grasping_task_robot}
            \end{subfigure}
            % % % % 
            \caption{Experiment setup of the circular object grasping task. }
            % % % % 
            \label{fig:grasping_task_setup}
        \end{figure}

        In this task the goal is to give an object with a round opening to the robot. The setup is depicted in Figure \ref{fig:grasping_task_setup}. The opening is modeled as a circle, i.e. the robot can grasp the object all around its opening. However, two additional constraints on the orientation are necessary to model this task. These constraints are shown in Figure \ref{fig:grasping_task_constraints}. The end-effector is required to be perpendicular to the plane that the circle lies in. A plane in geometric algebra can be obtained from a circle by a multiplying the circle with $\gae{\infty}$ and the normal vector is then obtained as 
        \begin{equation}\label{eq:ga_plane_normal}
            \bm{n}_E = E^* - 0.5 (E^* \inner \gae{0}) \gae{\infty}.
        \end{equation}
        The second constraint is that the direction that the gripper is actuated in needs to be perpendicular to the circle, which expressed mathematically means that this direction needs to be coaxial with the line that connects the grasping position and the center of the circle when it is projected into the plane of the circle. The projection of a point $P$ to the plane $E$ is computed as
        \begin{equation}\label{eq:ga_point_to_plane_projection}
            P' = (E\inner P)E^{-1}.
        \end{equation}

        % % % %

        \begin{figure}[!ht]
            % % % %
            \centering
            % % % %
            \includegraphics[width=\linewidth]{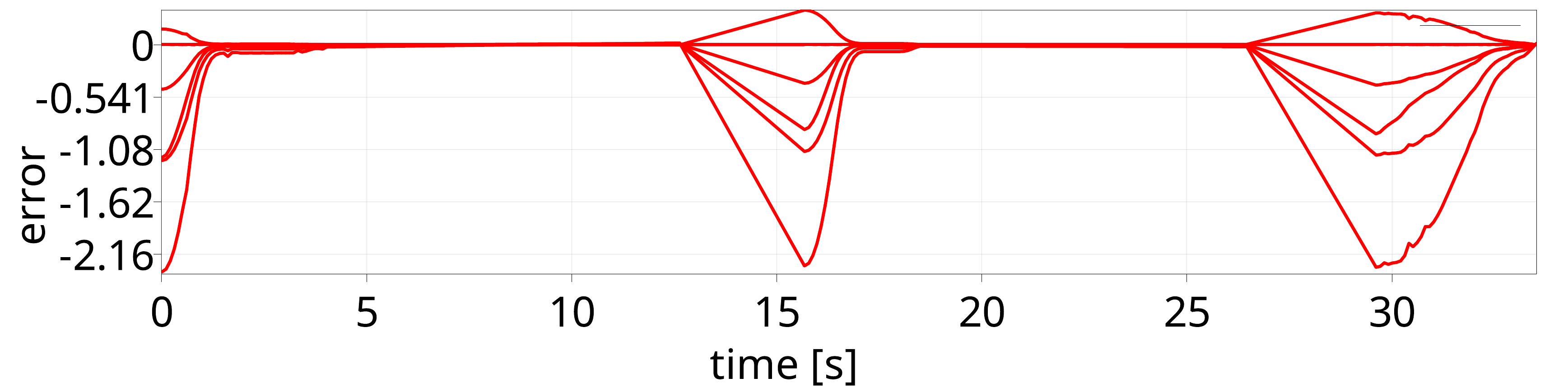}
            % % % % 
            \caption{Experimental results of the circular object grasping task. In this figure three repetitions are depicted, the location of the target box was changed in-between.}
            % % % % 
            \label{fig:grasping_task_results}
        \end{figure}

        In Figure \ref{fig:grasping_task_setup} we show the setup of the experiment with the Franka Emika robot. The box that we used has a circular opening and we defined it by measuring three points relative to an Aruco marker that we attached to the side of the box. Then during the experiment the robot was reaching for the box while satisfying the aforementioned constraints using MPC. As soon as it reached (i.e. the cost was below a threshold) the robot closed its gripper in order to hold the box. We repeated this experiment multiple times and changed the box position by holding in our hands for giving it to the robot. An excerpt of the resulting error vector over time can be found in Figure \ref{fig:grasping_task_results}.
    % subsection object_grasping_tasks (end) 

    % % % % % % % % % % % % % % % % % % % % % % % % % % % % % % % % % % % % % % % % % %
    
    % \subsection{Weight Compensation}
    % \label{sub:weight_compensation}
    %     A second task that serves the verification of the computed joint torques from Section \ref{sub:dynamics_of_serial_manipulators} is a simple weight compensation task. For this task we let the robot grasp various objects with a known weight and apply joint torques that compensate the gravitational forces of the objects. This experiment is run for 10s and in Figure \ref{fig:weight_compensation_results_} we show the deviation from the original position during this time. Note that it is not a feedback controller, we only apply feed-forward torques.

    %     % % % % 
    %     \begin{figure}[!ht]
    %         % % % %
    %         \centering
    %         % % % %
    %         \includegraphics[width=0.25\linewidth]{images/panda_weight_compensation.jpg}
    %         % % % %
    %         \caption{Weight compensation results.}
    %         % % % %
    %         \label{fig:weight_compensation_results_}
    %     \end{figure}
    % subsection sub:weight_compensation (end)
% section sec:experiments (end)

% % % % % % % % % % % % % % % % % % % % % % % % % % % % % % % % % % % % % % % % % % % % 

\section{Conclusion}
\label{sec:conclusion}
    We presented in this paper the usage of geometric algebra for the modelling of optimal control tasks and how to use the dynamics of serial manipulators computed with geometric algebra for inverse dynamics control.

    The provided library, \textit{gafro}, is currently specialized for conformal geometric algebra. The implementation of the multivectors and expressions that define the algebra is generic. Thus it would be possible to use geometric algebras with different signatures, which can be used to explore the usage of different geometric primitives such as quadric surfaces in this optimal control framework.

    Higher order quadric surfaces such as cones and paraboloids in $\mathbb{G}_{6,3}$ \cite{zamora-esquivelGeometricAlgebraDescription2014} or ellipsoids and hyperboloids in $\qcga$ \cite{breuilsQuadricConformalGeometric} are still a topic of ongoing research. In theory it should be possible to use them seamlessly in combination with the methods that we presented in this paper, since the properties of the different geometric algebras such as the outer product nullspace, which we rely on, remain the same. It is therefore the topic of future work to investigate the integration of these algebras into our formulation. The benefit of this would be a more versatile and generic modeling of surfaces that can be exploited for various manipulation tasks.

    A possible extension and application of this work would be grasping and in-hand manipulation. Using the geometric representations and the corresponding optimization functions presented in this we can actually very easily derive a model for grasping in geometric algebra. The three contact types point, line and plane can directly be represented as geometric primitives. In most cases the contact points are surface points of objects. Especially the most commonly used point-on-plane model, which is always stable.
% section sec:conclusion (end)

% \bibliographystyle{IEEEtran}
% \bibliography{references}
\printbibliography

@article{adornoDQRoboticsLibrary2021,
  title = {{{DQ Robotics}}: {{A Library}} for {{Robot Modeling}} and {{Control}}},
  shorttitle = {{{DQ Robotics}}},
  author = {Adorno, Bruno Vilhena and Marques Marinho, Murilo},
  date = {2021-09},
  journaltitle = {IEEE Robotics Automation Magazine},
  volume = {28},
  number = {3},
  pages = {102--116},
  issn = {1558-223X},
  doi = {10.1109/MRA.2020.2997920},
  eventtitle = {{{IEEE Robotics Automation Magazine}}}
}

@thesis{adornoTwoarmManipulationManipulators2011,
  type = {phdthesis},
  title = {Two-Arm {{Manipulation}}: {{From Manipulators}} to {{Enhanced Human-Robot Collaboration}}},
  author = {Adorno, Bruno Vilhena},
  date = {2011},
  institution = {{Université Montpellier II - Sciences et Techniques du Languedoc}},
  address = {Department of Robotics},
  langid = {english}
}

@article{afonsosilvaDynamicsMobileManipulators2022,
  title = {Dynamics of {{Mobile Manipulators Using Dual Quaternion Algebra}}},
  author = {Afonso Silva, Frederico Fernandes and José Quiroz-Omaña, Juan and Vilhena Adorno, Bruno},
  date = {2022-09-14},
  journaltitle = {Journal of Mechanisms and Robotics},
  volume = {14},
  number = {6},
  issn = {1942-4302},
  doi = {10.1115/1.4054320},
  urldate = {2022-06-12}
}

@article{arellano-muroNewtonEulerModeling2020,
  title = {Newton–{{Euler Modeling}} and {{Control}} of a {{Multi-copter Using Motor Algebra}}},
  author = {Arellano-Muro, Carlos A. and Osuna-González, Guillermo and Castillo-Toledo, Bernardino and Bayro-Corrochano, Eduardo},
  date = {2020-04},
  journaltitle = {Adv. Appl. Clifford Algebras},
  volume = {30},
  number = {2},
  pages = {19},
  issn = {0188-7009, 1661-4909},
  doi = {10.1007/s00006-020-1045-1},
  urldate = {2021-05-24},
  langid = {english}
}

@article{aristidouExtendingFABRIKModel2016,
  title = {Extending {{FABRIK}} with Model Constraints},
  author = {Aristidou, Andreas and Chrysanthou, Yiorgos and Lasenby, Joan},
  date = {2016-01},
  journaltitle = {Comp. Anim. Virtual Worlds},
  volume = {27},
  number = {1},
  pages = {35--57},
  issn = {1546-4261, 1546-427X},
  doi = {10.1002/cav.1630},
  urldate = {2021-05-24},
  langid = {english}
}

@incollection{aristidouInverseKinematicsSolutions2011,
  title = {Inverse {{Kinematics Solutions Using Conformal Geometric Algebra}}},
  booktitle = {Guide to {{Geometric Algebra}} in {{Practice}}},
  author = {Aristidou, Andreas and Lasenby, Joan},
  editor = {Dorst, Leo and Lasenby, Joan},
  date = {2011},
  pages = {47--62},
  publisher = {{Springer London}},
  location = {{London}},
  doi = {10.1007/978-0-85729-811-9_3},
  urldate = {2021-05-24},
  isbn = {978-0-85729-810-2 978-0-85729-811-9},
  langid = {english}
}

@article{bayro-corrochanoComputingConformalSpace2022,
  title = {Computing in the {{Conformal Space Objects}}, {{Incidence Relations}}, and {{Geometric Constrains}} for {{Applications}} in {{AI}}, {{GIS}}, {{Graphics}}, {{Robotics}}, and {{Human-Machine Interaction}}},
  author = {Bayro-Corrochano, Eduardo Jose and Altamirano-Escobedo, G. and Ortiz-Gonzalez, A. and Farias-Moreno, V. and Chel-Puc, N.},
  date = {2022},
  journaltitle = {IEEE Access},
  volume = {10},
  pages = {112742--112756},
  issn = {2169-3536},
  doi = {10.1109/ACCESS.2022.3216266},
  eventtitle = {{{IEEE Access}}}
}

@article{bayro-corrochanoDifferentialInverseKinematics2007,
  title = {Differential and Inverse Kinematics of Robot Devices Using Conformal Geometric Algebra},
  author = {Bayro-Corrochano, Eduardo and Zamora-Esquivel, Julio},
  date = {2007-01},
  journaltitle = {Robotica},
  volume = {25},
  number = {1},
  pages = {43--61},
  issn = {0263-5747, 1469-8668},
  doi = {10.1017/S0263574706002980},
  urldate = {2021-05-24},
  langid = {english}
}

@book{bayro-corrochanoGeometricAlgebraApplications2019,
  title = {Geometric {{Algebra Applications Vol}}. {{I}}},
  author = {Bayro-Corrochano, Eduardo},
  date = {2019},
  publisher = {{Springer International Publishing}},
  location = {{Cham}},
  doi = {10.1007/978-3-319-74830-6},
  urldate = {2022-04-09},
  isbn = {978-3-319-74828-3 978-3-319-74830-6},
  langid = {english}
}

@book{bayro-corrochanoGeometricAlgebraApplications2020,
  title = {Geometric {{Algebra Applications Vol}}. {{II}}: {{Robot Modelling}} and {{Control}}},
  shorttitle = {Geometric {{Algebra Applications Vol}}. {{II}}},
  author = {Bayro-Corrochano, Eduardo},
  date = {2020},
  publisher = {{Springer International Publishing}},
  location = {{Cham}},
  doi = {10.1007/978-3-030-34978-3},
  urldate = {2021-05-27},
  isbn = {978-3-030-34976-9 978-3-030-34978-3},
  langid = {english}
}

@article{bayro-corrochanoGeometricIntuitiveTechniques2020,
  title = {Geometric {{Intuitive Techniques}} for {{Human Machine Interaction}} in {{Medical Robotics}}},
  author = {Bayro-Corrochano, Eduardo and Garza-Burgos, Ana M. and Del-Valle-Padilla, Juan L.},
  date = {2020-01},
  journaltitle = {Int J of Soc Robotics},
  volume = {12},
  number = {1},
  pages = {91--112},
  issn = {1875-4791, 1875-4805},
  doi = {10.1007/s12369-019-00545-8},
  urldate = {2021-05-24},
  langid = {english}
}

@article{bayro-corrochanoMotorAlgebraApproach2000,
  title = {Motor Algebra Approach for Computing the Kinematics of Robot Manipulators},
  author = {Bayro-Corrochano, Eduardo and Kähler, Detlef},
  date = {2000},
  journaltitle = {Journal of Robotic Systems},
  volume = {17},
  number = {9},
  pages = {495--516},
  issn = {1097-4563},
  doi = {10.1002/1097-4563(200009)17:9<495::AID-ROB4>3.0.CO;2-S},
  urldate = {2021-06-15},
  langid = {english}
}

@article{bayro-corrochanoNewtonEulerModeling2022,
  title = {Newton–{{Euler}} Modeling and {{Hamiltonians}} for Robot Control in the Geometric Algebra},
  author = {Bayro-Corrochano, Eduardo and Medrano-Hermosillo, Jesus and Osuna-González, Guillermo and Uriostegui-Legorreta, Ulises},
  date = {2022-11},
  journaltitle = {Robotica},
  volume = {40},
  number = {11},
  pages = {4031--4055},
  publisher = {{Cambridge University Press}},
  issn = {0263-5747, 1469-8668},
  doi = {10.1017/S0263574722000741},
  urldate = {2023-03-14},
  langid = {english}
}

@article{bayro-corrochanoSurveyQuaternionAlgebra2021,
  title = {A {{Survey}} on {{Quaternion Algebra}} and {{Geometric Algebra Applications}} in {{Engineering}} and {{Computer Science}} 1995–2020},
  author = {Bayro-Corrochano, Eduardo},
  date = {2021},
  journaltitle = {IEEE Access},
  volume = {9},
  pages = {104326--104355},
  issn = {2169-3536},
  doi = {10.1109/ACCESS.2021.3097756},
  eventtitle = {{{IEEE Access}}}
}

@inproceedings{beesonTRACIKOpensourceLibrary2015,
  title = {{{TRAC-IK}}: {{An}} Open-Source Library for Improved Solving of Generic Inverse Kinematics},
  shorttitle = {{{TRAC-IK}}},
  booktitle = {2015 {{IEEE-RAS}} 15th {{International Conference}} on {{Humanoid Robots}} ({{Humanoids}})},
  author = {Beeson, Patrick and Ames, Barrett},
  date = {2015-11},
  pages = {928--935},
  publisher = {{IEEE}},
  location = {{Seoul, South Korea}},
  doi = {10.1109/HUMANOIDS.2015.7363472},
  urldate = {2023-03-09},
  eventtitle = {2015 {{IEEE-RAS}} 15th {{International Conference}} on {{Humanoid Robots}} ({{Humanoids}})},
  isbn = {978-1-4799-6885-5},
  langid = {english}
}

@inproceedings{belonApplicationsConformalGeometric2013,
  title = {Applications of {{Conformal Geometric Algebra}} in {{Mesh Deformation}}},
  booktitle = {2013 {{XXVI Conference}} on {{Graphics}}, {{Patterns}} and {{Images}}},
  author = {Belon, Mauricio Cele Lopez},
  date = {2013-08},
  pages = {39--46},
  publisher = {{IEEE}},
  location = {{Arequipa, Peru}},
  doi = {10.1109/SIBGRAPI.2013.15},
  urldate = {2021-05-29},
  eventtitle = {2013 {{XXVI SIBGRAPI}} - {{Conference}} on {{Graphics}}, {{Patterns}} and {{Images}} ({{SIBGRAPI}})},
  isbn = {978-0-7695-5099-2},
  langid = {english}
}

@article{breuilsGaramonGeometricAlgebra2019,
  title = {Garamon: {{A Geometric Algebra Library Generator}}},
  shorttitle = {Garamon},
  author = {Breuils, Stéphane and Nozick, Vincent and Fuchs, Laurent},
  date = {2019-07-22},
  journaltitle = {Adv. Appl. Clifford Algebras},
  volume = {29},
  number = {4},
  pages = {69},
  issn = {1661-4909},
  doi = {10.1007/s00006-019-0987-7},
  urldate = {2022-01-13},
  langid = {english}
}

@article{breuilsNewApplicationsClifford2022,
  title = {New {{Applications}} of {{Clifford}}’s {{Geometric Algebra}}},
  author = {Breuils, Stephane and Tachibana, Kanta and Hitzer, Eckhard},
  date = {2022-04},
  journaltitle = {Adv. Appl. Clifford Algebras},
  volume = {32},
  number = {2},
  pages = {17},
  issn = {0188-7009, 1661-4909},
  doi = {10.1007/s00006-021-01196-7},
  urldate = {2022-04-20},
  langid = {english}
}

@article{breuilsQuadricConformalGeometric,
  title = {Quadric {{Conformal Geometric Algebra}} of {{R9}},6},
  author = {Breuils, Stephane and Nozick, Vincent and Hitzer, Eckhard},
  pages = {15},
  langid = {english}
}

@article{calinonGaussiansRiemannianManifolds2020,
  title = {Gaussians on {{Riemannian Manifolds}}: {{Applications}} for {{Robot Learning}} and {{Adaptive Control}}},
  shorttitle = {Gaussians on {{Riemannian Manifolds}}},
  author = {Calinon, Sylvain},
  date = {2020-06},
  journaltitle = {IEEE Robotics Automation Magazine},
  volume = {27},
  number = {2},
  pages = {33--45},
  issn = {1558-223X},
  doi = {10.1109/MRA.2020.2980548},
  eventtitle = {{{IEEE Robotics Automation Magazine}}}
}

@inproceedings{carpentierPinocchioLibraryFast2019,
  title = {The {{Pinocchio C}}++ Library : {{A}} Fast and Flexible Implementation of Rigid Body Dynamics Algorithms and Their Analytical Derivatives},
  shorttitle = {The {{Pinocchio C}}++ Library},
  booktitle = {2019 {{IEEE}}/{{SICE International Symposium}} on {{System Integration}} ({{SII}})},
  author = {Carpentier, Justin and Saurel, Guilhem and Buondonno, Gabriele and Mirabel, Joseph and Lamiraux, Florent and Stasse, Olivier and Mansard, Nicolas},
  date = {2019-01},
  pages = {614--619},
  publisher = {{IEEE}},
  location = {{Paris, France}},
  doi = {10.1109/SII.2019.8700380},
  urldate = {2022-10-25},
  eventtitle = {2019 {{IEEE}}/{{SICE International Symposium}} on {{System Integration}} ({{SII}})},
  isbn = {978-1-5386-3615-2}
}

@thesis{colapinto2011versor,
  title = {Versor: {{Spatial}} Computing with Conformal Geometric Algebra},
  author = {Colapinto, Pablo},
  date = {2011},
  institution = {{University of California at Santa Barbara}},
  url = {http://versor.mat.ucsb.edu},
  key = {colapinto2011thesis}
}

@article{dantamRobustEfficientForward2021,
  title = {Robust and Efficient Forward, Differential, and Inverse Kinematics Using Dual Quaternions},
  author = {Dantam, Neil T},
  date = {2021-09},
  journaltitle = {The International Journal of Robotics Research},
  volume = {40},
  number = {10-11},
  pages = {1087--1105},
  issn = {0278-3649, 1741-3176},
  doi = {10.1177/0278364920931948},
  urldate = {2023-03-08},
  langid = {english}
}

@book{doranGeometricAlgebraPhysicists2003,
  title = {Geometric {{Algebra}} for {{Physicists}}},
  author = {Doran, Chris and Lasenby, Anthony},
  date = {2003-05-29},
  edition = {1},
  publisher = {{Cambridge University Press}},
  doi = {10.1017/CBO9780511807497},
  urldate = {2022-06-01},
  isbn = {978-0-521-48022-2 978-0-521-71595-9 978-0-511-80749-7},
  langid = {english}
}

@incollection{fernandesExploringLazyEvaluation2021,
  title = {Exploring {{Lazy Evaluation}} and {{Compile-Time Simplifications}} for {{Efficient Geometric Algebra Computations}}},
  booktitle = {Systems, {{Patterns}} and {{Data Engineering}} with {{Geometric Calculi}}},
  author = {Fernandes, Leandro A. F.},
  editor = {Xambó-Descamps, Sebastià},
  date = {2021},
  volume = {13},
  pages = {111--131},
  publisher = {{Springer International Publishing}},
  location = {{Cham}},
  doi = {10.1007/978-3-030-74486-1_6},
  urldate = {2022-11-07},
  isbn = {978-3-030-74485-4 978-3-030-74486-1},
  langid = {english}
}

@article{fonsecaCoupledTaskSpaceAdmittance2020,
  title = {Coupled {{Task-Space Admittance Controller Using Dual Quaternion Logarithmic Mapping}}},
  author = {Fonseca, Mariana de Paula Assis and Adorno, Bruno Vilhena and Fraisse, Philippe},
  date = {2020-10},
  journaltitle = {IEEE Robotics and Automation Letters},
  volume = {5},
  number = {4},
  pages = {6057--6064},
  issn = {2377-3766},
  doi = {10.1109/LRA.2020.3010458},
  eventtitle = {{{IEEE Robotics}} and {{Automation Letters}}}
}

@inproceedings{fontijneGaigenGeometricAlgebra2006,
  title = {Gaigen 2:: A Geometric Algebra Implementation Generator},
  shorttitle = {Gaigen 2},
  booktitle = {Proceedings of the 5th International Conference on {{Generative}} Programming and Component Engineering  - {{GPCE}} '06},
  author = {Fontijne, Daniel},
  date = {2006},
  pages = {141},
  publisher = {{ACM Press}},
  location = {{Portland, Oregon, USA}},
  doi = {10.1145/1173706.1173728},
  urldate = {2022-11-07},
  eventtitle = {The 5th International Conference},
  isbn = {978-1-59593-237-2},
  langid = {english}
}

@article{gonzalez-jimenezRobustPoseControl2014,
  title = {Robust {{Pose Control}} of {{Robot Manipulators Using Conformal Geometric Algebra}}},
  author = {González-Jiménez, L. and Carbajal-Espinosa, O. and Loukianov, A. and Bayro-Corrochano, E.},
  date = {2014-06},
  journaltitle = {Adv. Appl. Clifford Algebras},
  volume = {24},
  number = {2},
  pages = {533--552},
  issn = {0188-7009, 1661-4909},
  doi = {10.1007/s00006-014-0448-2},
  urldate = {2021-05-24},
  langid = {english}
}

@article{gunnGeometricAlgebrasEuclidean2017,
  title = {Geometric {{Algebras}} for {{Euclidean Geometry}}},
  author = {Gunn, Charles},
  date = {2017-03},
  journaltitle = {Adv. Appl. Clifford Algebras},
  volume = {27},
  number = {1},
  pages = {185--208},
  issn = {0188-7009, 1661-4909},
  doi = {10.1007/s00006-016-0647-0},
  urldate = {2021-05-24},
  langid = {english}
}

@incollection{hadfieldConstrainedDynamicsConformal2020,
  title = {Constrained {{Dynamics}} in {{Conformal}} and {{Projective Geometric Algebra}}},
  booktitle = {Advances in {{Computer Graphics}}},
  author = {Hadfield, Hugo and Lasenby, Joan},
  editor = {Magnenat-Thalmann, Nadia and Stephanidis, Constantine and Wu, Enhua and Thalmann, Daniel and Sheng, Bin and Kim, Jinman and Papagiannakis, George and Gavrilova, Marina},
  date = {2020},
  volume = {12221},
  pages = {459--471},
  publisher = {{Springer International Publishing}},
  location = {{Cham}},
  doi = {10.1007/978-3-030-61864-3_39},
  urldate = {2021-05-24},
  isbn = {978-3-030-61863-6 978-3-030-61864-3},
  langid = {english}
}

@article{hadfieldExploringNovelSurface2021,
  title = {Exploring {{Novel Surface Representations}} via an {{Experimental Ray-Tracer}} in {{CGA}}},
  author = {Hadfield, Hugo and Achawal, Sushant and Lasenby, Joan and Lasenby, Anthony and Young, Benjamin},
  date = {2021-04},
  journaltitle = {Adv. Appl. Clifford Algebras},
  volume = {31},
  number = {2},
  pages = {16},
  issn = {0188-7009, 1661-4909},
  doi = {10.1007/s00006-021-01117-8},
  urldate = {2021-05-24},
  langid = {english}
}

@article{heNovelAdaptiveFiltering2020,
  title = {Novel {{Adaptive Filtering Algorithms Based}} on {{Higher-Order Statistics}} and {{Geometric Algebra}}},
  author = {He, Yinmei and Wang, Rui and Wang, Xiangyang and Zhou, Jian and Yan, Yi},
  date = {2020},
  journaltitle = {IEEE Access},
  volume = {8},
  pages = {73767--73779},
  issn = {2169-3536},
  doi = {10.1109/ACCESS.2020.2988521},
  urldate = {2021-05-24},
  langid = {english}
}

@book{hestenesCliffordAlgebraGeometric1984,
  title = {Clifford {{Algebra}} to {{Geometric Calculus}}},
  author = {Hestenes, David and Sobczyk, Garret},
  date = {1984},
  publisher = {{Springer Netherlands}},
  location = {{Dordrecht}},
  doi = {10.1007/978-94-009-6292-7},
  urldate = {2021-05-27},
  isbn = {978-90-277-2561-5 978-94-009-6292-7},
  langid = {english}
}

@article{hestenesCliffordAlgebraGeometric1985,
  title = {\emph{Clifford }{{\emph{Algebra}}}\emph{ to }{{\emph{Geometric Calculus}}}\emph{. }{{\emph{A Unified Language}}}\emph{ for} {{\emph{Mathematics}}}\emph{ and }{{\emph{Physics}}}},
  author = {Hestenes, David and Sobczyk, Garret and Marsh, James S.},
  date = {1985-05},
  journaltitle = {American Journal of Physics},
  volume = {53},
  number = {5},
  pages = {510--511},
  issn = {0002-9505, 1943-2909},
  doi = {10.1119/1.14223},
  urldate = {2021-06-02},
  langid = {english}
}

@article{hitzerCurrentSurveyClifford2022,
  title = {Current Survey of {{Clifford}} Geometric Algebra Applications},
  author = {Hitzer, Eckhard and Lavor, Carlile and Hildenbrand, Dietmar},
  date = {2022-04-29},
  journaltitle = {Math Methods in App Sciences},
  pages = {mma.8316},
  issn = {0170-4214, 1099-1476},
  doi = {10.1002/mma.8316},
  urldate = {2022-05-19},
  langid = {english}
}

@article{hwangboPerContactIterationMethod2018,
  title = {Per-{{Contact Iteration Method}} for {{Solving Contact Dynamics}}},
  author = {Hwangbo, Jemin and Lee, Joonho and Hutter, Marco},
  date = {2018-04},
  journaltitle = {IEEE Robot. Autom. Lett.},
  volume = {3},
  number = {2},
  pages = {895--902},
  issn = {2377-3766, 2377-3774},
  doi = {10.1109/LRA.2018.2792536},
  urldate = {2022-06-20},
  langid = {english}
}

@article{jaquierGeometryawareManipulabilityLearning2021,
  title = {Geometry-Aware Manipulability Learning, Tracking, and Transfer},
  author = {Jaquier, Noémie and Rozo, Leonel and Caldwell, Darwin G and Calinon, Sylvain},
  date = {2021-02},
  journaltitle = {The International Journal of Robotics Research},
  volume = {40},
  number = {2-3},
  pages = {624--650},
  issn = {0278-3649, 1741-3176},
  doi = {10.1177/0278364920946815},
  urldate = {2021-05-24},
  langid = {english}
}

@software{jeremy_ong_2019,
  title = {{{GAL}}},
  author = {Ong, Jeremy},
  date = {2019},
  location = {{https://github.com/jeremyong/gal}},
  url = {https://github.com/jeremyong/gal},
  commit = {[insert commit used]},
  organization = {{GitHub}}
}

@article{johansenDualQuaternionControl2019,
  title = {Dual Quaternion Control: A Review of Recent Results within Motion Control},
  author = {Johansen, Tor-Aleksander and Sanchez, José J Corona and Kristiansen, Raymond},
  date = {2019},
  journaltitle = {Nonlinear Studies},
  volume = {26},
  number = {4},
  pages = {25},
  langid = {english}
}

@book{jootGeometricAlgebraElectrical2019,
  title = {Geometric {{Algebra}} for {{Electrical Engineers}}},
  author = {Joot, Peeter},
  date = {2019},
  publisher = {{CreateSpace Independent Publishing Platform}},
  langid = {english}
}

@article{kamarianakisAllInOneGeometricAlgorithm2021,
  title = {An {{All-In-One Geometric Algorithm}} for {{Cutting}}, {{Tearing}}, and {{Drilling Deformable Models}}},
  author = {Kamarianakis, Manos and Papagiannakis, George},
  date = {2021-07-06},
  journaltitle = {Adv. Appl. Clifford Algebras},
  number = {31},
  eprint = {2102.07499},
  eprinttype = {arxiv},
  eprintclass = {cs},
  url = {http://arxiv.org/abs/2102.07499},
  urldate = {2022-11-22},
  langid = {english}
}

@report{kavanDualQuaternionsRigid2006,
  type = {Technical Report},
  title = {Dual {{Quaternions}} for {{Rigid Transformation Blending}}},
  author = {Kavan, Ladislav and Collins, Steven and O’Sullivan, Carol and Zara, Jiri},
  date = {2006},
  pages = {11},
  institution = {{Trinity College}},
  location = {{Dublin, Ireland}},
  langid = {english}
}

@article{kolpashchikovFABRIKxTacklingInverse2022,
  title = {{{FABRIKx}}: {{Tackling}} the {{Inverse Kinematics Problem}} of {{Continuum Robots}} with {{Variable Curvature}}},
  shorttitle = {{{FABRIKx}}},
  author = {Kolpashchikov, Dmitrii and Gerget, Olga and Danilov, Viacheslav},
  date = {2022-12},
  journaltitle = {Robotics},
  volume = {11},
  number = {6},
  pages = {128},
  publisher = {{Multidisciplinary Digital Publishing Institute}},
  issn = {2218-6581},
  doi = {10.3390/robotics11060128},
  urldate = {2022-11-30},
  issue = {6},
  langid = {english}
}

@article{lechuga-gutierrezIterativeInverseKinematics2022,
  title = {Iterative Inverse Kinematics for Robot Manipulators Using Quaternion Algebra and Conformal Geometric Algebra},
  author = {Lechuga-Gutierrez, L. and Macias-Garcia, E. and Martínez-Terán, G. and Zamora-Esquivel, J. and Bayro-Corrochano, E.},
  date = {2022-06},
  journaltitle = {Meccanica},
  volume = {57},
  number = {6},
  pages = {1413--1428},
  issn = {0025-6455, 1572-9648},
  doi = {10.1007/s11012-022-01512-w},
  urldate = {2023-03-07},
  langid = {english}
}

@article{leclercq3DKinematicsUsing2013,
  title = {{{3D}} Kinematics Using Dual Quaternions: Theory and Applications in Neuroscience},
  shorttitle = {{{3D}} Kinematics Using Dual Quaternions},
  author = {Leclercq, Guillaume and Lefèvre, Philippe and Blohm, Gunnar},
  date = {2013},
  journaltitle = {Front. Behav. Neurosci.},
  volume = {7},
  issn = {1662-5153},
  doi = {10.3389/fnbeh.2013.00007},
  urldate = {2022-04-20},
  langid = {english}
}

@article{lopesGeometricAlgebraAdaptiveFilters2019,
  title = {Geometric-{{Algebra Adaptive Filters}}},
  author = {Lopes, Wilder Bezerra and Lopes, Cassio Guimaraes},
  date = {2019-07-15},
  journaltitle = {IEEE Trans. Signal Process.},
  volume = {67},
  number = {14},
  pages = {3649--3662},
  issn = {1053-587X, 1941-0476},
  doi = {10.1109/TSP.2019.2916028},
  urldate = {2021-05-24},
  langid = {english}
}

@article{macdonaldSurveyGeometricAlgebra2017,
  title = {A {{Survey}} of {{Geometric Algebra}} and {{Geometric Calculus}}},
  author = {Macdonald, Alan},
  date = {2017-03},
  journaltitle = {Adv. Appl. Clifford Algebras},
  volume = {27},
  number = {1},
  pages = {853--891},
  issn = {0188-7009, 1661-4909},
  doi = {10.1007/s00006-016-0665-y},
  urldate = {2021-05-24},
  langid = {english}
}

@article{maricRiemannianMetricGeometryaware2021,
  title = {A {{Riemannian}} Metric for Geometry-Aware Singularity Avoidance by Articulated Robots},
  author = {Marić, Filip and Petrović, Luka and Guberina, Marko and Kelly, Jonathan and Petrović, Ivan},
  date = {2021-11},
  journaltitle = {Robotics and Autonomous Systems},
  volume = {145},
  pages = {103865},
  issn = {09218890},
  doi = {10.1016/j.robot.2021.103865},
  urldate = {2022-03-07},
  langid = {english}
}

@article{maricRiemannianOptimizationDistanceGeometric2022,
  title = {Riemannian {{Optimization}} for {{Distance-Geometric Inverse Kinematics}}},
  author = {Marić, Filip and Giamou, Matthew and Hall, Adam W. and Khoubyarian, Soroush and Petrovic, Ivan and Kelly, Jonathan},
  date = {2022-06},
  journaltitle = {IEEE Trans. Robot.},
  volume = {38},
  number = {3},
  pages = {1703--1722},
  issn = {1552-3098, 1941-0468},
  doi = {10.1109/TRO.2021.3123841},
  urldate = {2022-06-20},
  langid = {english}
}

@inproceedings{marinhoDualQuaternionLinearquadratic2015,
  title = {A Dual Quaternion Linear-Quadratic Optimal Controller for Trajectory Tracking},
  booktitle = {2015 {{IEEE}}/{{RSJ International Conference}} on {{Intelligent Robots}} and {{Systems}} ({{IROS}})},
  author = {Marinho, M. M. and Figueredo, L. F. C. and Adorno, B. V.},
  date = {2015-09},
  pages = {4047--4052},
  doi = {10.1109/IROS.2015.7353948},
  eventtitle = {2015 {{IEEE}}/{{RSJ International Conference}} on {{Intelligent Robots}} and {{Systems}} ({{IROS}})}
}

@article{marinhoDynamicActiveConstraints2019,
  title = {Dynamic {{Active Constraints}} for {{Surgical Robots Using Vector-Field Inequalities}}},
  author = {Marinho, Murilo Marques and Adorno, Bruno Vilhena and Harada, Kanako and Mitsuishi, Mamoru},
  date = {2019-10},
  journaltitle = {IEEE Trans. Robot.},
  volume = {35},
  number = {5},
  pages = {1166--1185},
  issn = {1552-3098, 1941-0468},
  doi = {10.1109/TRO.2019.2920078},
  urldate = {2022-07-18},
  langid = {english}
}

@book{perwassGeometricAlgebraApplications2009,
  title = {Geometric Algebra with Applications in Engineering},
  author = {Perwass, Christian},
  date = {2009},
  series = {Geometry and Computing},
  number = {4},
  publisher = {{Springer}},
  location = {{Berlin}},
  isbn = {978-3-540-89067-6},
  langid = {english},
  pagetotal = {385}
}

@book{sicilianoRobotics2009,
  title = {Robotics},
  author = {Siciliano, Bruno and Sciavicco, Lorenzo and Villani, Luigi and Oriolo, Giuseppe},
  editorb = {Grimble, Michael J. and Johnson, Michael A.},
  editorbtype = {redactor},
  date = {2009},
  series = {Advanced {{Textbooks}} in {{Control}} and {{Signal Processing}}},
  publisher = {{Springer London}},
  location = {{London}},
  doi = {10.1007/978-1-84628-642-1},
  urldate = {2021-12-07},
  isbn = {978-1-84628-641-4 978-1-84628-642-1},
  langid = {english}
}

@software{smitsKDLKinematicsDynamics,
  title = {{{KDL}}: {{Kinematics}} and {{Dynamics Library}}},
  author = {Smits, R},
  location = {{http://www.orocos.org/kdl}},
  url = {http://www.orocos.org/kdl},
  urldate = {25.10.2022}
}

@article{sousaTbGALTensorBasedLibrary2020,
  title = {{{TbGAL}}: {{A Tensor-Based Library}} for {{Geometric Algebra}}},
  shorttitle = {{{TbGAL}}},
  author = {Sousa, Eduardo Vera and Fernandes, Leandro A. F.},
  date = {2020-04},
  journaltitle = {Adv. Appl. Clifford Algebras},
  volume = {30},
  number = {2},
  pages = {27},
  issn = {0188-7009, 1661-4909},
  doi = {10.1007/s00006-020-1053-1},
  urldate = {2021-12-05},
  langid = {english}
}

@inproceedings{tassaSynthesisStabilizationComplex2012,
  title = {Synthesis and Stabilization of Complex Behaviors through Online Trajectory Optimization},
  booktitle = {2012 {{IEEE}}/{{RSJ International Conference}} on {{Intelligent Robots}} and {{Systems}}},
  author = {Tassa, Yuval and Erez, Tom and Todorov, Emanuel},
  date = {2012-10},
  pages = {4906--4913},
  issn = {2153-0866},
  doi = {10.1109/IROS.2012.6386025},
  eventtitle = {2012 {{IEEE}}/{{RSJ International Conference}} on {{Intelligent Robots}} and {{Systems}}}
}

@thesis{tingelstadAutomaticDifferentiationOptimization2017,
  type = {phdthesis},
  title = {Automatic {{Differentiation}} and {{Optimization}} of {{Multivectors}}: {{Estimating Motors}} in {{Conformal Geometric Algebra}}},
  shorttitle = {Automatic {{Differentiation}} and {{Optimization}} of {{Multivectors}}},
  author = {Tingelstad, Lars},
  date = {2017},
  institution = {{NTNU}},
  address = {Department of Mechanical and Industrial Engineering},
  url = {http://rgdoi.net/10.13140/RG.2.2.10157.10722},
  urldate = {2021-05-24},
  langid = {english}
}

@article{tingelstadMotorParameterization2018,
  title = {Motor {{Parameterization}}},
  author = {Tingelstad, Lars and Egeland, Olav},
  date = {2018-05},
  journaltitle = {Adv. Appl. Clifford Algebras},
  volume = {28},
  number = {2},
  pages = {34},
  issn = {0188-7009, 1661-4909},
  doi = {10.1007/s00006-018-0850-2},
  urldate = {2021-06-07},
  langid = {english}
}

@article{zamora-esquivelGeometricAlgebraDescription2014,
  title = {G 6,3 {{Geometric Algebra}}; {{Description}} and {{Implementation}}},
  author = {Zamora-Esquivel, Julio},
  date = {2014-06},
  journaltitle = {Adv. Appl. Clifford Algebras},
  volume = {24},
  number = {2},
  pages = {493--514},
  issn = {0188-7009, 1661-4909},
  doi = {10.1007/s00006-014-0442-8},
  urldate = {2021-06-10},
  langid = {english}
}

@article{zamora-esquivelRobotObjectManipulation2011,
  title = {Robot {{Object Manipulation Using Stereoscopic Vision}} and {{Conformal Geometric Algebra}}},
  author = {Zamora-Esquivel, Julio and Bayro-Corrochano, Eduardo},
  date = {2011},
  journaltitle = {Applied Bionics and Biomechanics},
  volume = {8},
  number = {3-4},
  pages = {411--428},
  issn = {1176-2322, 1754-2103},
  doi = {10.1155/2011/728132},
  urldate = {2021-11-09},
  langid = {english}
}

@article{zaplanaClosedformSolutionsInverse2022,
  title = {Closed-Form Solutions for the Inverse Kinematics of Serial Robots Using Conformal Geometric Algebra},
  author = {Zaplana, Isiah and Hadfield, Hugo and Lasenby, Joan},
  date = {2022},
  journaltitle = {Mechanism and Machine Theory},
  volume = {173},
  eprint = {2109.12411},
  eprinttype = {arxiv},
  doi = {10.1016/j.mechmachtheory.2022.104835},
  langid = {english}
}

% % % % % % % % % % % % % % % % % % % % % % % % % % % % % % % % % % % % % % % % % % % % 

\appendix
    % % % % % % % % % % % % % % % % % % % % % % % % % % % % % % % % % % % % % % % % % %
    
    \subsection{Structure of a Multivector in Conformal Geometric Algebra}
    \label{sub:structure_of_a_multivector_in_conformal_geometric_algebra}
        Conformal geometric algebra has 32 basis blades, following the rule $k=2^{r+p+q}$. Grade 5 is the highest grade of CGA, consequently that makes $\gae{0123\infty}$ the pseudoscalar of the algebra. In practice the multivectors are sparse and usually at most 10 elements are non-zero. We call the vector containing the known non-zero blades of a multivector its parameter vector. Note that the grade 0 element is a scalar which means that vectors and matrices known from linear algebra are actually part of this algebra and can therefore be seamlessly multiplied with multivector matrices.
        \begin{table}[!ht]
            \scriptsize
            \centering
            % % % % 
            \begin{tabular}{ll}
                \toprule
                    grade 0
                    & 1
                    \\
                \midrule
                    grade 1
                    & $\gae{1}$, $\gae{2}$, $\gae{3}$, $\gae{\infty}$, $\gae{0}$
                    \\
                \midrule
                    grade 2
                    & $\gae{23}$, $\gae{13}$, $\gae{12}$, $\gae{1\infty}$, $\gae{2\infty}$, $\gae{3\infty}$, $\gae{01}$, $\gae{02}$, $\gae{03}$, $\gae{0\infty}$
                    \\
                \midrule
                    grade 3
                    & $\gae{123}$, $\gae{12\infty}$, $\gae{13\infty}$, $\gae{23\infty}$, $\gae{012}$, $\gae{013}$, $\gae{023}$, $\gae{01\infty}$, $\gae{02\infty}$, $\gae{03\infty}$
                    \\
                \midrule
                    grade 4
                    & $\gae{123\infty}$, $\gae{0123}$, $\gae{012\infty}$, $\gae{023\infty}$, $\gae{013\infty}$
                    \\
                \midrule
                    grade 5
                    & $\gae{0123\infty}$
                    \\
                \bottomrule
            \end{tabular}
            \caption{Basis blades of conformal geometric algebra.}
            \label{tab:basis_blades_of_conformal_geometric_algebra}
        \end{table}
    % subsection structure_of_a_multivector_in_conformal_geometric_algebra (end)
    
    \subsection{Geometric Primitives in Conformal Geometric Algebra}
    \label{sec:geometric_primitives_in_conformal_geometric_algebra}
        Table \ref{tab:geometric_primitives_in_cga} shows the geometric primitives that are available in conformal geometric algebra. We define the outer product nullspace as the primal representation, because of its more convenient construction of the primitives. Note that all primitives can be used as the argument of the sandwich product $MX\reverse{M}$, i.e. applying rigid body transformations to the primitives. Higher dimensional algebras such as $\qcga$ extend these primitives by quadric surfaces such as ellipsoids and hyperboloids.
        % % % % 
        \begin{table}[!ht]
            % % % %
            \centering
            \cgaprimitives
            \vspace*{1mm}
            % % % %
            \caption{Geometric primitives in CGA \cite{perwassGeometricAlgebraApplications2009}}
            % % % %
            \label{tab:geometric_primitives_in_cga}
        \end{table}
        There are several geometric operations that allow for reasoning about the relationships between primitives such as projections and intersections. The latter can actually be found as well in the construction of the geometric primitives, for example a circle in its IPNS representation is constructed as the outer product of two spheres, i.e. the intersection of two spheres. 

    \subsection{Embedding}
    \label{sec:embedding}
        This section explains the embedding function $\mathcal{E}\left[ \gamatrix{X} \right]$ that is used to obtain the non-trivial parameter matrix of a multivector matrix. Suppose you have an arbitrary multivector matrix $\gamatrix{X} \in \mathbb{R}^{I\times J}$ with elements $X_{ij}$ that have $K$ known non-trivial blades $\gae{k}$. The resulting parameter matrix then is of size $IK\times J$, i.e. the parameter vectors of each multivector element get expanded along the columns of the matrix. The embedding function $\mathcal{E}$ is therefore defined as
        \begin{equation}\label{eq:ga_embedding_function}
            \bm{X} =  \mathcal{E}\left[ \gamatrix{X} \right] = \mat{
                X_{11,1}
                & \hdots
                & X_{1J,1}
                \\    
                \vdots
                & \ddots
                & \vdots
                \\    
                X_{11,K}
                & \hdots
                & X_{1J,1}
                \\    
                \vdots
                & \ddots
                & \vdots
                \\    
                X_{I1,K}
                & \hdots
                & X_{IJ,1}
                \\    
                \vdots
                & \ddots
                & \vdots
                \\    
                X_{I1,K}
                & \hdots
                & X_{IJ,K}
            }.
        \end{equation}
        \textcolor{black}{Note that the embedding function can be considered an implementation detail that assures a minimal memory consumption and easy integration with off-the-shelf optimization solvers using matrix representations. Therefore it is handled automatically by the library without the user having to set it explicitly. }
    % % % % % % % % % % % % % % % % % % % % % % % % % % % % % % % % % % % % % % % % % %
    
    \subsection{Derivation of the Jacobian of the Motor Logarithmic Map}
    \label{sub:derivation_of_the_jacobian_of_the_motor_logarithmic_map}
        From Equation \eqref{eq:motor_exp_log_map} we know that $B = \log(M)$. We define the parameters of the motor and bivector as follows
        \begin{multline}\label{eq:motor_parameters}
            M = m_1 + m_2 \gae{23} + m_3 \gae{13} + m_4 \gae{12} \\ + m_5 \gae{1\infty} + m_6 \gae{2\infty} + m_7 \gae{3\infty} + m_8 \gae{123\infty},
        \end{multline}
        and
        \begin{equation}\label{eq:bivector_parameters}
            B = b_1 \gae{23} + b_2 \gae{13} + b_3 \gae{12} + b_4 \gae{1\infty} + b_5 \gae{2\infty} + b_6 \gae{3\infty}.
        \end{equation}
        Standard results show that the motor $M$ can be split into a rotor $R$ and a translator $T$, such that $M = TR$
        \begin{eqnarray}
            R & = & -\gae{0} \cdot M \gae{\infty},
            \\
            T & = & M \reverse{R}.
        \end{eqnarray}
        Using the Equations \eqref{eq:cgarotor} and \eqref{eq:cgatranslator} it becomes straightforward to derive the bivector components $b_i$ in function of the motor components $m_i$   
        \begin{eqnarray}\label{eq:bivector_from_motor_components}
            b_1 &=& -m_2 \frac{2\acos{m_1}}{\sin\left( \acos {m_1} \right) },
            \\
            b_2 &=& -m_3 \frac{2\acos{m_1}}{\sin\left( \acos {m_1} \right) },
            \\
            b_3 &=& -m_4 \frac{2\acos{m_1}}{\sin\left( \acos {m_1} \right) },
            \\
            b_4 &=& -2(m_1m_5 + m_4m_6 + m_3m_7 + m_2m_8),
            \\
            b_5 &=& -2(-m_4m_5 + m_1m_6 + m_2m_7 - m_3m_8),
            \\
            b_6 &=& -2(-m_3m_5 - m_2m_6 + m_1m_7 + m_4m_8).
        \end{eqnarray}

        The Jacobian of the motor logarithmic map can be found as the partial derivatives of the bivector components $b_i$ w.r.t the motor components $m_i$, i.e.
        \begin{equation}
            \bm{J}_{\mathcal{M}\to\mathbb{B}}(M) = \mat{
                \frac{\partial b_1}{\partial m_1} & \ldots & \frac{\partial b_1}{\partial m_8}
                \\
                \vdots & \ddots & \vdots
                \\
                \frac{\partial b_6}{\partial m_1} & \ldots & \frac{\partial b_6}{\partial m_8}
            }.
        \end{equation}
        Using the Equations \eqref{eq:bivector_from_motor_components} the non-trivial partial derivatives can be found to be
        \begin{eqnarray}
           \frac{\partial b_1}{\partial m_1} = -2m_2 \left( \frac{1}{m_1^2-1} + m_1 \acos{m_1} \right) ,
           \\
           \frac{\partial b_2}{\partial m_1} = -2m_3 \left( \frac{1}{m_1^2-1} + m_1 \acos{m_1} \right) ,
           \\
           \frac{\partial b_3}{\partial m_1} = -2m_4 \left( \frac{1}{m_1^2-1} + m_1 \acos{m_1} \right) ,
           \\
           \frac{\partial b_1}{\partial m_2} = \frac{\partial b_2}{\partial m_3} = \frac{\partial b_3}{\partial m_4} = \frac{-2\acos{m_1}}{\sin\left( \acos{m_1} \right) },
           \\
           \frac{\partial b_4}{\partial m_1} = -\frac{\partial b_5}{\partial m_4} = -\frac{\partial b_6}{\partial m_3}= -2m_5,
           \\
           \frac{\partial b_4}{\partial m_2} = -\frac{\partial b_5}{\partial m_3} = \frac{\partial b_6}{\partial m_4}  = -2m_8,
           \\
           \frac{\partial b_4}{\partial m_3} = \frac{\partial b_5}{\partial m_2} = \frac{\partial b_6}{\partial m_1} = -2m_7,
           \\
           \frac{\partial b_4}{\partial m_4} = \frac{\partial b_5}{\partial m_1} = -\frac{\partial b_6}{\partial m_2}  = -2m_6,
           \\
           \frac{\partial b_4}{\partial m_5} = \frac{\partial b_5}{\partial m_6} = \frac{\partial b_6}{\partial m_7} = -2m_1,
           \\
           \frac{\partial b_4}{\partial m_6} = -\frac{\partial b_5}{\partial m_5} = \frac{\partial b_6}{\partial m_8}= -2m_4,
           \\
           \frac{\partial b_4}{\partial m_7} = -\frac{\partial b_5}{\partial m_8} = -\frac{\partial b_6}{\partial m_5} = -2m_3,
           \\
           \frac{\partial b_4}{\partial m_8} = \frac{\partial b_5}{\partial m_7} = -\frac{\partial b_6}{\partial m_6}= -2m_2,
       \end{eqnarray}
       which concludes the derivation.                                     
        
    % subsection derivation_of_the_jacobian_of_the_motor_logarithmic_map (end)
% appendix (end)
        
% % % % % % % % % % % % % % % % % % % % % % % % % % % % % % % % % % % % % % % % % % % % 

\begin{IEEEbiography}[{\includegraphics[width=1in,height=1.25in,clip,keepaspectratio]{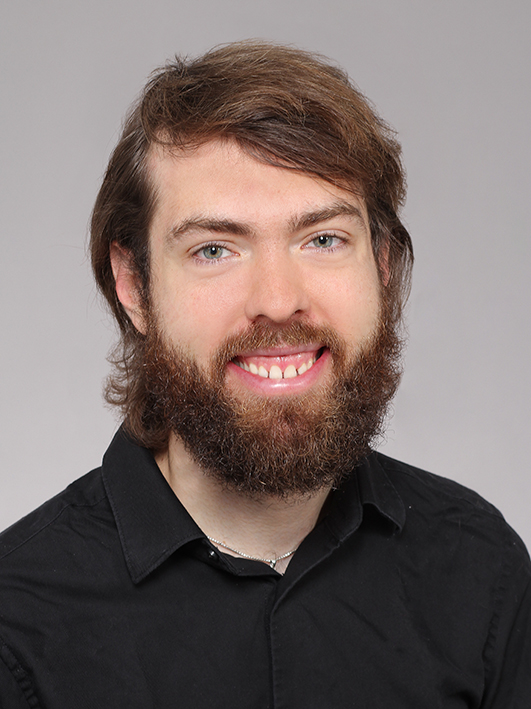}}]{Tobias L\"ow} is a Ph.D. student in Electrical Engineering at École Polytechnique Fédérale de Lausanne (EPFL) and is working as a research assistant in the Robot Learning and Interaction Group at the Idiap Research Institute. He received his BSc. and MSc. in Mechanical Engineering from ETH Z\"urich, Switzerland, in 2018 and 2020, respectively. He conducted his master's thesis at the Robotics and Autonomous Systems Group, CSIRO, Brisbane. His research interests lie in exploiting geometric and structural methods for optimization problems in robotics. \newline Website: \url{https://tobiloew.ch}.
\end{IEEEbiography}

\begin{IEEEbiography}[{\includegraphics[width=1in,height=1.25in,clip,keepaspectratio]{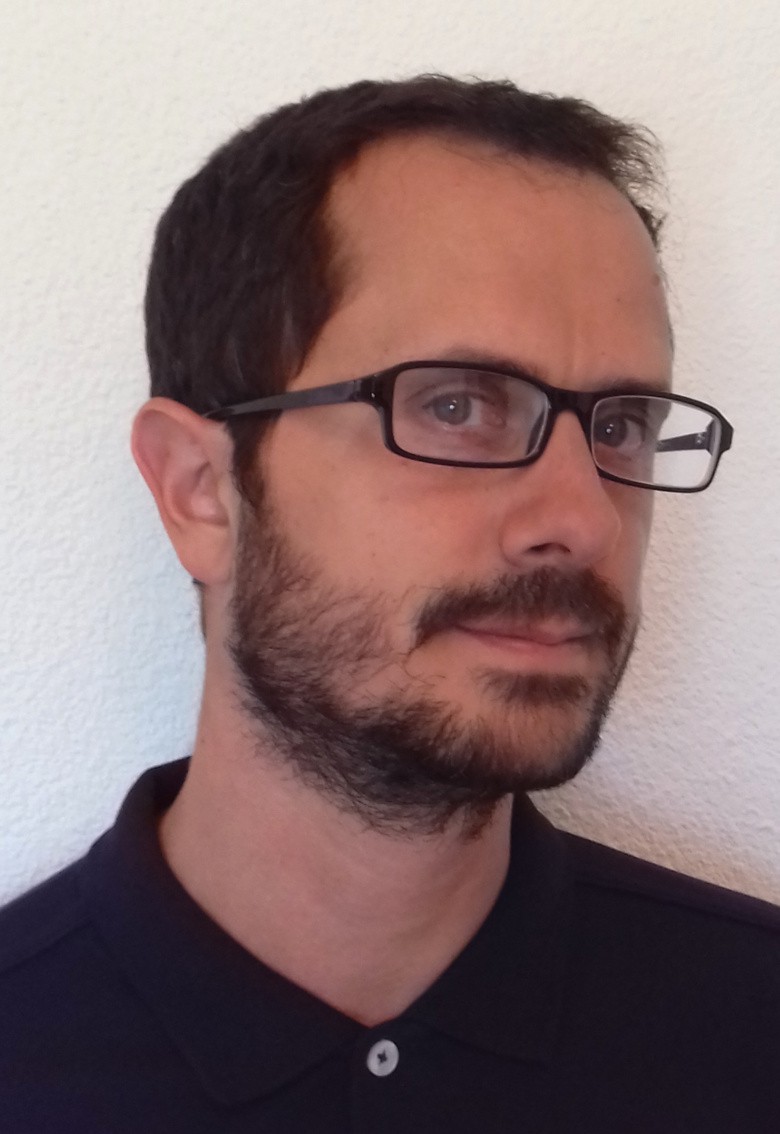}}]{Sylvain Calinon} received the Ph.D. degree in robotics from the École Polytechnique Fedérale de Lausanne (EPFL), Lausanne, Switzerland, in 2007. He is currently a Senior Researcher with the Idiap Research Institute, Martigny, Switzerland, and a Lecturer with the EPFL. From 2009 to 2014, he was a Team Leader with the Italian Institute of Technology, Genoa, Italy. From 2007 to 2009, he was a Postdoc with EPFL. His research interests include human–robot collaboration, robot learning, and model-based optimization. Website: \url{https://calinon.ch}.
\end{IEEEbiography}

\end{document}